\pdfoutput=1

\documentclass[11pt]{article}

\usepackage{acl}

\usepackage{times}
\usepackage{latexsym}

\usepackage[T1]{fontenc}

\usepackage[utf8]{inputenc}

\usepackage{microtype}

\usepackage{graphicx}
\usepackage{amsmath}
\usepackage{amssymb}
\usepackage{booktabs}
\usepackage[font=normalsize]{caption}
\usepackage{textcomp}

\usepackage{subcaption}
\usepackage{xcolor}
\usepackage{algorithm}
\usepackage{algpseudocode}
\usepackage{booktabs}

\usepackage[capitalize]{cleveref}
\crefname{section}{Sec.}{Secs.}
\Crefname{section}{Section}{Sections}
\Crefname{table}{Table}{Tables}
\crefname{table}{Tab.}{Tabs.}

\title{NL2TL: Transforming Natural Languages to Temporal Logics using Large Language Models}

\author{Yongchao Chen \\
  MIT / Harvard \\
  \small\texttt{ycchen98@mit.edu} \\\And
  Rujul Gandhi \\
  MIT \\
  \small\texttt{rujul@mit.edu} \\\And
  Yang Zhang \\
  MIT-IBM Watson AI Lab \\
  \small\texttt{Yang.Zhang2@ibm.com} \\\And
  Chuchu Fan \\
  MIT \\
  \small\texttt{chuchu@mit.edu} \\  }

\begin{document}
\maketitle
\begin{abstract}
Temporal Logic (TL) can be used to rigorously specify complex high-level specification for systems in many engineering applications. The translation between natural language (NL) and TL has been under-explored due to the lack of dataset and generalizable model across different application domains.  
In this paper, we propose an accurate and generalizable transformation framework of English instructions from NL to TL, exploring the use of Large Language Models (LLMs) at multiple stages. Our contributions are twofold. First, we develop a framework to create a dataset of NL-TL pairs combining LLMs and human annotation. We publish a dataset with 28K NL-TL pairs. Then, we finetune T5 models on the lifted versions (i.e., the specific Atomic Propositions (AP) are hidden) of the NL and TL. The enhanced generalizability originates from two aspects: 1) Usage of lifted NL-TL characterizes common logical structures, without constraints of specific domains. 2) Application of LLMs in dataset creation largely enhances corpus richness. We test the generalization of trained models on five varied domains. To achieve full NL-TL transformation, we either combine the lifted model with AP recognition task or do the further finetuning on each specific domain. During the further finetuning, our model achieves higher accuracy (>95\%) using only <10\% training data, compared with the baseline sequence to sequence (Seq2Seq) model.\footnote{Datasets and Codes are available at \url{https://github.com/yongchao98/NL2TL}}\footnote{Project Page is available at \url{https://yongchao98.github.io/MIT-realm-NL2TL}}
\end{abstract}

\section{Introduction}
Temporal Logic (TL) has been widely used as a mathematically precise language to specify requirements in many engineering domains such as robotics \cite{tellex2020robots}, electronics design \cite{browne1986automatic}, autonomous driving \cite{9304549}. TL can capture the complex spatial, temporal, and logical requirements of both human languages and environmental constraints, and can be transformed into executable actions or control inputs for robots \cite{gundana2022event,raman2013sorry,boteanu2016model,patel2020grounding,gopalan2018sequence}.

Unlike many robotics works that try to use end-to-end black-box models to infer robotic behaviors directly from natural language (NL) \cite{google2022saycan}, using structured TL as the intermediate has a twofold benefit -- the TL can be used for direct planning, and the TL representation can be used to identify specific sources of failure and provide automatic feedback to a non-expert user \cite{raman2013sorry}. However, TL has a steep learning curve. Communicating one's goals and constraints through NL is much more intuitive to a non-expert. Therefore, a model able to transform NL instructions into TL is a missing but crucial component for interactive robots and engineering designs.

Currently, there is no general tool to perform automated translations between TL and NL that takes the following requirements into consideration:

\begin{itemize}
    \item \textbf{Cross-domain generalization.} Although TL is used in many engineering domains, current NL-to-TL approaches largely constrain their training data to a single domain. These datasets mostly lack plentiful corpus richness of NL-TL and have their own specified formats of Atomic Propositions (AP). Then the models fail to generalize to other domains \cite{gopalan2018sequence}, even though the structure of TL itself is not dependent on the domain and should be generic. 
    \item \textbf{Variability of NL instructions.} Past work often constructs synthetic data algorithmically, requiring limited forms of the NL input. Real-world NL utterances cannot be encoded into a small set of rules. Models trained on such homogeneous data fail to generalize to new sentence structures \cite{brunello2019synthesis}.
    
\end{itemize}

One big bottleneck in the NL-to-TL problem is the lack of data. Although modern statistical methods can outperform rule-based methods \cite{buzhinsky2019survey}, they typically require a huge dataset. This data is expensive and difficult to collect since strong expertise of annotators is needed \cite{brunello2019synthesis}. As outlined above, constraining the domain or form of the NL instructions relieves the pressure of dataset, but also unavoidably undermines the generalizability \cite{brunello2019synthesis,patel2019learning}.

To supplement the data creation process and simultaneously overcome the need for a huge dataset, we propose to use pre-trained LLMs. We utilize GPT-3 \cite{brown2020language} to assist dataset creation and finetune T5 models \cite{raffel2020exploring} to be specialized in NL-to-TL transformation.

Another aspect of our approach is to use `lifted' versions of NL and TL for finetuning our model, which greatly enhances generalizability. In previous work, models trained on full NL-to-TL transformation often include converting specific individual actions into APs. For example, the AP \emph{"a response is created in Slack"} might be formalized as \emph{"create\_Slack"}. As a result, each work has to regularize its own content and style of APs, affecting generalization. Instead of taking this approach, we hide all the APs in our data during finetuning, acquiring a lifted model on lifted NL-to-TL transformation. For the final ground application from full NL into full TL, two methods are proposed, either by combining the lifted model with AP recognition or further transfer learning in one specific domain. For further transfer learning into specific domains, we compare the models with/without pre-training on lifted NL-TL and show its significance.

In this paper, we present two key contributions:
\begin{itemize}
  \item \textbf{Constructing a cross-domian NL-TL dataset.} We generate a dataset of 15K lifted NL-TL pairs using a novel GPT-3-assisted framework. Ablation studies are conducted to show the significance of each part of the framework for dataset construction. In addition, we collect and clean previous datasets (13K) from two varied domains, adapting original full NL-TL pairs into lifted versions. In this way, we publish a dataset of 28K lifted NL-TL pairs. Ablation studies show that the newly created data are indispensable since purely training on the collected data fails to work across domains.
  \item \textbf{Finetuning a lifted NL-to-TL model} on T5 using our data, and demonstrating the improvement in performance compared to former state-of-the-art methods. For application in full NL-to-STL transformation, two methods are proposed. We compare our model to Seq2Seq models and direct few-shot learning by GPT-3, across five domains. The experimental results show that our methods achieve better accuracy (>95\% across all domains) and are more data-efficient (<10\% domain specific data). We also do the ablation study by training a Seq2Seq model with lifted NL-to-TL dataset, revealing that T5's superior model capacity is essential.
\end{itemize}

GPT-4 \cite{bubeck2023sparks} comes out when approaching the end of this work. To compare the accuracy of direct GPT-4 few-shot end-to-end translation with our finetuned model, we did an ad-hoc test on ChatGPT Plus version with 100 samples in each domain. Here we can not test on more samples since we do not have access to GPT-4 API and ChatGPT Plus version has access limitation per hour. The results show that GPT-4 achieves an accuracy of 77.7\% over 300 samples, much lower than our model but higher than GPT-3.

\begin{figure}[t]
  \centering
  \includegraphics[width=0.95\linewidth]{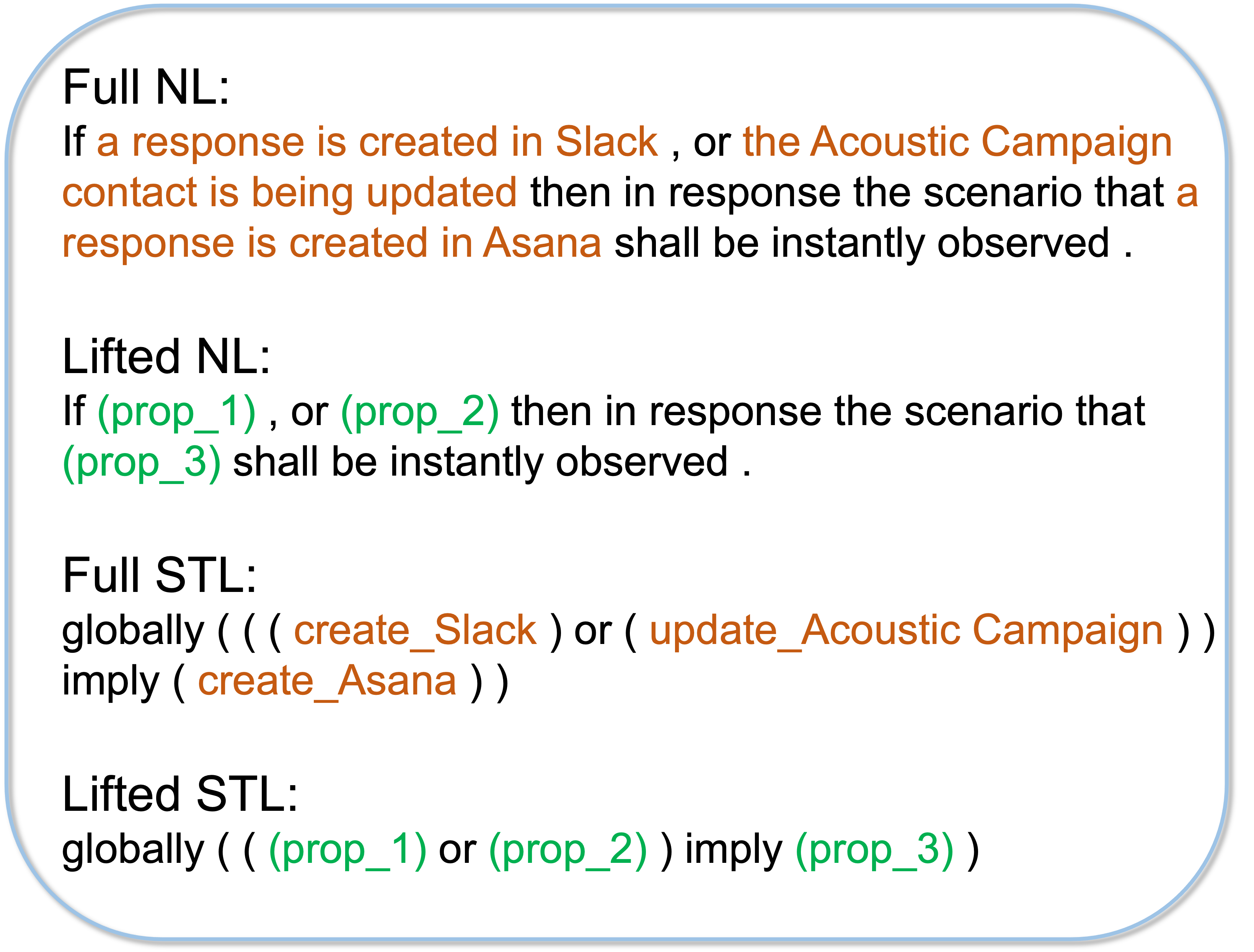}
   \caption{
Illustration of lifted NL and lifted STL.}
   \label{fig:onefigure}
\end{figure}

\section{Temporal Logic Specifications}
\subsection{STL Syntax}
There are many different versions of TL \cite{emerson1990temporal,maler2004monitoring,koymans1990specifying}. They are more or less similar in terms of syntax. We will use Signal Temporal Logic (STL) as a representative formal language that supports constraints over the continuous real-time, which is more suitable to capture time-critical missions. In some previous work, Linear Temporal Logic (LTL) is also widely used, which is contained by STL when the time is discrete.  We will construct our framework based on STL and show that the trained model also performs well on datasets and tasks using LTL. An STL formula is defined recursively according to the following syntax:
\begin{multline}
\phi ::= \pi^\mu\;|\;\neg\phi\;|\;\phi\land\varphi\;|\;\phi\lor\varphi\;|\;\textbf{F}_{[a,b]}\phi\;|\;\textbf{G}_{[a,b]}\phi\;\\
|\;\phi\textbf{U}_{[a,b]}\varphi
\end{multline}
where $\phi$ and $\varphi$ are STL formulas, and $\pi^\mu$ is an atomic predicate. $\neg$ (negation), $\land$ (and), $\lor$ (or), $\Rightarrow$ (imply), and $\Leftrightarrow$ (equal)) are logical operators. $\textbf{F}_{[a,b]}$ (eventually/finally), $\textbf{G}_{[a,b]}$ (always/globally), and $\textbf{U}_{[a,b]}$ (until) are temporal operators with real-time constraints $t \in [a, b]$. Temporal operators with time constraints are illustrated by Table~\ref{tab:STL}, and other operators can be presented using the basic syntax.

\subsection{Lifted STL and Lifted NL}
We represent our data as `lifted' NL and STL, in which the specific APs corresponding to individual actions are hidden (following nomenclature from \citet{hsiung2021generalizing}). In our lifted NL and STL, each AP is replaced with a placeholder \emph{prop\_i}. In this way, we train our model on the general context of the instruction regardless of the specific APs. The correspondences between full and lifted NL/STL are shown in Figure~\ref{fig:onefigure}.

\subsection{STL Expression Formats}
\label{sec:STL formats}
\begin{figure*}[t]
  \centering
  \includegraphics[width=0.9\linewidth]{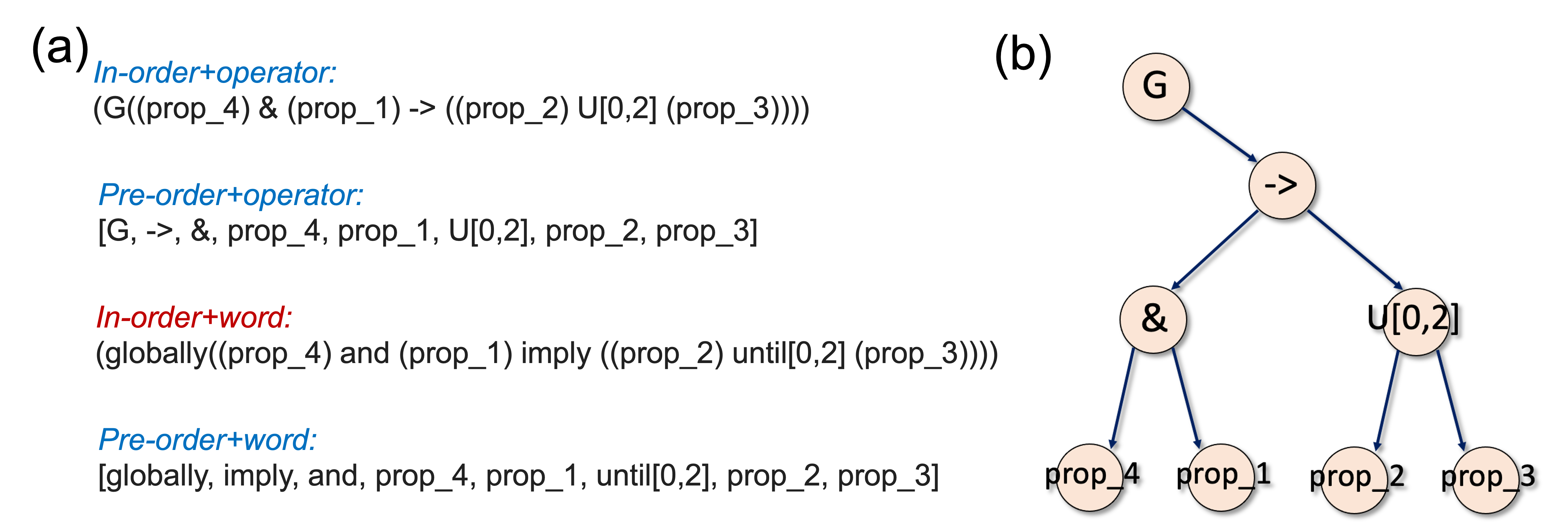}
   \caption{
Illustration of different formats of STL expressions. (a) Different expression formats of the same STL. (b) The binary tree representation of STL.}
   \label{fig:fivefigure}
\end{figure*}

Consider an STL expression as a binary tree, as in Figure~\ref{fig:fivefigure}. When finetuning a text-to-text model, there are different ways of representing the target STL in the form of linear text. Specifically, the targeted tokens can be linearized in an in-order (left subtree, root, right subtree) or pre-order (root, left subtree, right subtree) manner. Meanwhile, the operators can also be represented as the words with their corresponding meanings (rather than as symbols). The training results show that the in-order expression with all the operators replaced by words achieves much better accuracy than other three forms (will discuss in the following Section~\ref{sec:experimental results}).

\section{Related Work}
Over decades, researchers have methods to translate English sentences into various TL formulae \cite{brunello2019synthesis,finucane2010ltlmop,tellex2020robots,raman2013sorry}. However, to simplify the tasks, some previous work typically make strong assumptions to restrict the input text or the output formula, thus limiting the flexibility and generalizability.

The first representative attempt is by \citet{finucane2010ltlmop, tellex2011approaching, howard2014natural}, where the traditional methods typically follow three steps: 1) pre-process given English input by extracting syntactical information, 2) identify patterns or rules for TL through classification, and 3) run an attribute grammar-based parser to derive a target logical format. These methods only work for restricted input NL \cite{tellex2020robots}.

Another category of approaches are learning-based. Representative state-of-the-art works are \citet{gopalan2018sequence,wang2021learning,he2022deepstl}. In \citet{gopalan2018sequence} the authors gather a dataset focusing on Geometric LTL (GLTL), in which the NL and GLTL examples are all for the navigation of a car in the room. Then Seq2Seq models with attention mechanism are trained. Though the accuracy (93.45\%) is satisfying, the used GLTLs are relatively simple with each GLTL normally including one to three APs and the dataset also focuses on one confined task. In \citet{he2022deepstl} the authors choose to first translate a manually generated set of STL formulae into English sentences and train a semantic parser on the synthetic data. Such synthetic data cannot represent general NL and therefore the trained parser only works well on the original STL formulae. 

In \citet{wang2021learning} a semantic parser is built to learn the latent structure of NL commands for ground robots. The parser will provide (potentially incorrect) intermediate LTL representations to a motion planner, and the planner will give an executed trajectory as the feedback to see whether the robot’s execution satisfies the English input. Such approach has no guarantee on the correctness of the translated TL. In recent months, the work by \citet{aaai2023fc} directly applies LLMs like GPT-3 to convert NL to LTL via few-shot learning. The prompts should be well designed and the model will fail once the NL and LTL structures are too complex (we will discuss it in Section~\ref{subsection:AP recognition}).

Hence, in the domain of NL-to-TL translation, data augmentation/synthesis has been done algorithmically in previous work, not using generative models. This constrains how natural the resulting ‘NL’ actually is. In recent years, starting from the attention mechanism \cite{vaswani2017attention}, the rapid progression of pre-trained LLMs in NLP tends to unify many previous seemingly independent tasks into one large pre-trained model, especially the GPT series from OpenAI \cite{brown2020language}, and T5 \cite{raffel2020exploring} and PaLM \cite{chowdhery2022palm} from Google. These models are pre-trained with large amounts of natural sentences and codes, intrinsically encoding much world knowledge \cite{creswell2022selection}. The auto-regressive LLMs can naturally generate rich and meaningful corpus based on the given prompt. Then many recent work propose to do the data augmentation with LLMs, such as generating medical dialogues \cite{chintagunta2021medically} and python codes \cite{chen2021evaluating} via GPT-3. This inspires us the new opportunity in NL-to-TL task.

\section{Approach}
There are 3 steps in our approach. First, generating lifted NL-STL dataset with LLMs. Second, finetuning LLMs to get high accuracy on lifted NL-STL transformation. Third, lifting the data and applying the lifted model. Finally, we also consider the case where lifting is not possible and we must translate end to end by further finetuning the model.
\subsection{Data Generation}

\begin{algorithm}
\caption{Algorithm for STL synthesis}\label{alg:cap}
\begin{algorithmic}
 \renewcommand{\algorithmicrequire}{\textbf{Input:}}
  \renewcommand{\algorithmicensure}{\textbf{Output:}}
\Require 
{\\ Maximum number of APs $N$}
\Ensure {\\ Synthesized pre-order STL}
\\
\\ $two\_subtree$ = [$\land$, $\lor$, $\Rightarrow$, $\Leftrightarrow$, $\textbf{U}$, $\textbf{U}_{[a,b]}$]
\\ $one\_subtree$ = [$\neg$, $\textbf{F}$, $\textbf{G}$, $\textbf{F}_{[a,b]}$, $\textbf{G}_{[a,b]}$]
\\ sub\_lists $\gets$ Random prop\_list with total length [1, $N$] \textcolor{blue}{\Comment{[prop\_3, prop\_1], [prop\_2]}}

\\ Each sub\_list $\gets$ insert operators in $one\_subtree$ + $two\_subtree$ \textcolor{blue}{\Comment{[$\Leftrightarrow$, $\neg$, prop\_3, prop\_1], [$\textbf{G}$, prop\_2]}}

\\ Assembling sub\_lists into pre-order STL by appending random $two\_subtree$ operators
\textcolor{blue}{\Comment{[$\textbf{U}_{[10,30]}$, $\Leftrightarrow$, $\neg$, prop\_3, prop\_1, $\textbf{G}$, prop\_2]}}
\end{algorithmic}
\end{algorithm}

\begin{figure}[t]
  \centering
  \includegraphics[width=0.95\linewidth]{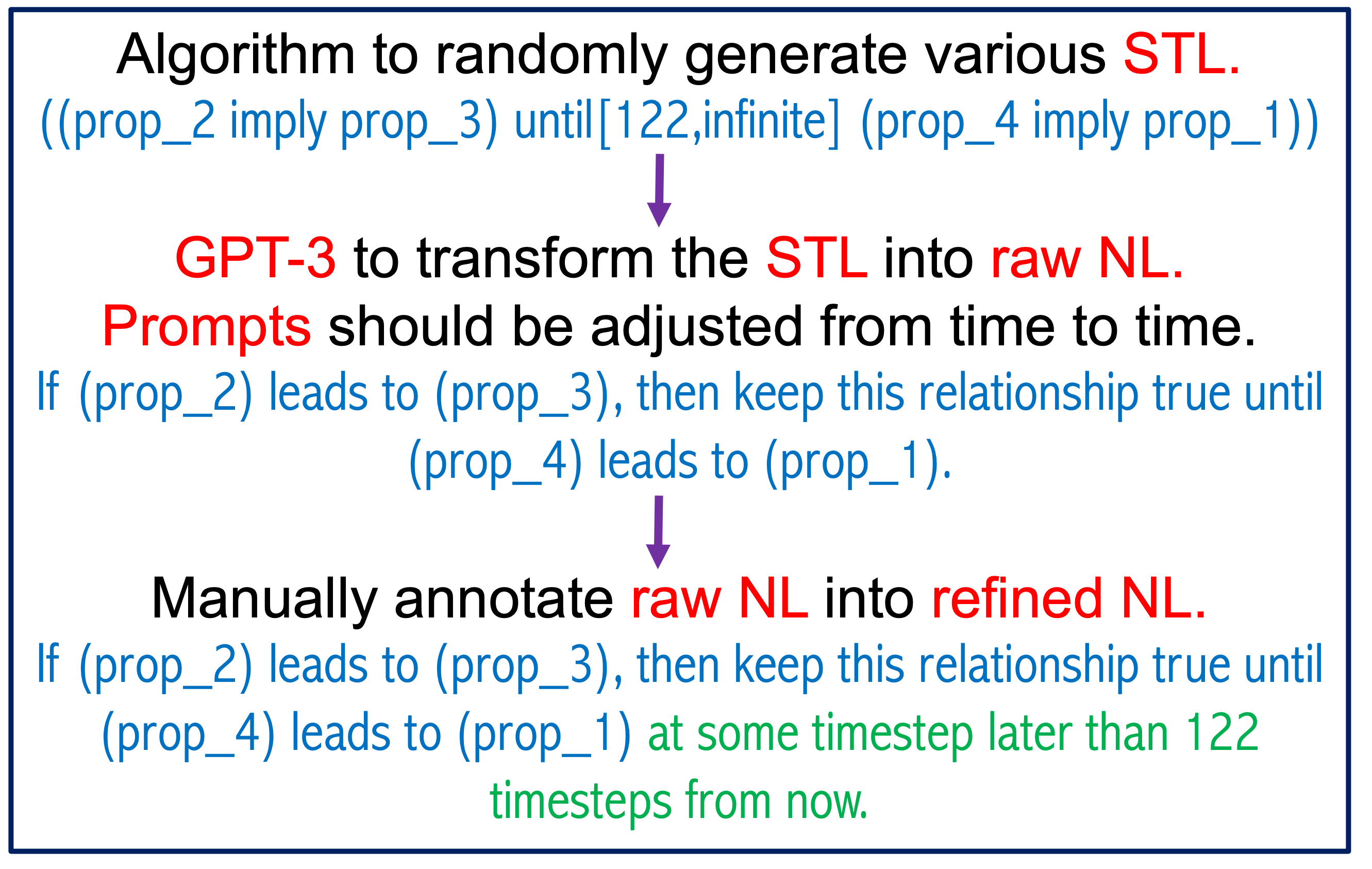}
   \caption{
Framework1 to generate NL-STL pairs.}
   \label{fig:twofigure}
\end{figure}

\begin{figure}[t]
  \centering
  \includegraphics[width=0.9\linewidth]{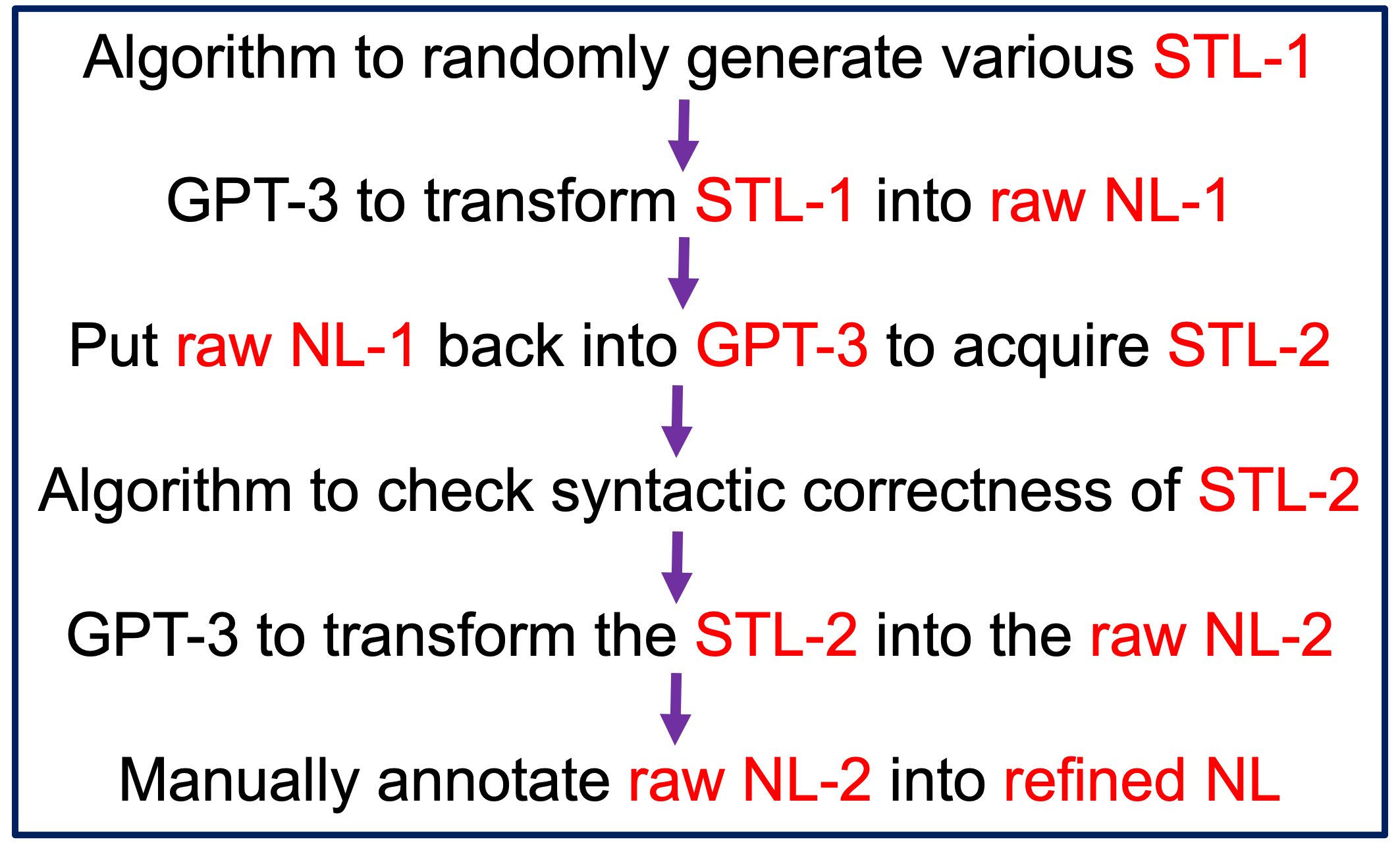}
   \caption{
Framework2 to generate NL-STL pairs. One extra loop between NL and STL is added.}
   \label{fig:threefigure}
\end{figure}
We apply the LLM GPT-3 (Davinci-003) to help generate multiple lifted NL and STL pairs. The first intuitive method is to use various NL-STL pairs as prompts and ask GPT-3 to automatically generate more NL-STL pairs. However, it turns out that the model always generates STL and NL with similar syntactic structures as the given prompts, thus limiting the sentence richness. To stimulate GPT-3 to generate sentences with more variations, we ask it to generate corresponding NLs from different STLs. The whole framework (referred as Framework1) is shown in Figure~\ref{fig:twofigure}. Various pre-order STLs are randomly synthesized by binary tree generation algorithm (See Algorithm~\ref{alg:cap} and specific discussion in Appendix~\ref{sec:appendixalgo}). The pre-order STLs are then transformed into in-order expressions via rules. To make the input STL more understandable to GPT-3, the operators are represented with the words of their meanings ($\Rightarrow(imply), \Leftrightarrow(equal), \lor(or), etc$). Then the GPT-3 will try to generate the raw NL whose semantic meaning is close to the STL. Human annotators then modify the raw NL to make its meaning consistent with the STL. During this process, the NL-STL pairs in prompts will be randomly chosen to make the vocabulary and sentence structure more diversified. We gather 200 NL instructions from 10 volunteers who are familiar with robot tasks and randomly choose 100 NL to serve as the prompt pool, while the other 100 NL serve as the Manual testing data. In each iteration of Framework1, 20 pairs are randomly chosen from the prompt pool to form the prompt of GPT-3 (a prompt example is shown in Appendix~\ref{sec:appendixA1} and the discussion on how many examples should be included in GPT-3 prompt is shown in Appendix~\ref{sec:appendix pair num vs acc}).

While Framework1 enhances the sentence richness, one issue is that the pure rule-based synthesis of STL sometimes generates unreasonable semantic meanings, or that the STL is too complex to describe it with NL. To solve this problem, an optimized framework (referred as Framework2) is shown in Figure~\ref{fig:threefigure}. Compared to Framework1, an extra loop between STL and NL is added using GPT-3. In this way, the initial rule-based STL with unreasonable or complex meanings will be automatically filtered by GPT-3 itself. In other words, during the mapping from STL-1 to NL-1, GPT-3 more or less modifies the meanings of the STLs that it cannot fully translate. Then the translated NL-1, though not fully consistent with STL-1, is more reasonable in the view of humans. It turns out that the semantic meanings synthesized by Framework2 are closer to the common human languages, and NL-STL pairs contain many fewer errors to annotate. The average number of annotated pairs is about 80 per person per hour with Framework1, and about 120 per person per hour with Framework2. 

We in total create 15108 lifted NL-STL pairs combining both Framework1 and Framework2, with the whole cost of around 150 person-hours. Appendix~\ref{sec:appendixA2} shows a prompt example to transform from NL-1 back into STL-2 via GPT-3, and Appendix~\ref{sec:appendixC} shows some example annotations of lifted NL-STL pairs. Appendix~\ref{sec:explanation for human annotation} explains the whole process and the license of human annotation to correct NL-STL pairs.

Apart from synthesizing and annotating lifted NL-STL pairs with GPT-3, we also collect and annotate the data gathered from \citet{wang2021learning} and \citet{he2022deepstl}. \citet{wang2021learning} focuses on robot navigation task with LTL, and \citet{he2022deepstl} focuses on circuit controlling with STL. To clean and process the data into lifted NL-STL pairs, the APs in both two datasets are detected and hidden by combining hard-coded algorithms with entity recognition package SpaCy \cite{honnibal2017spacy}. We gather 5K lifted NL-STL pairs from Navigation dataset \cite{wang2021learning} and 8K lifted NL-STL pairs from the Circuit dataset \cite{he2022deepstl}. Note that the original Navigation dataset uses LTL, while we correct some expression formats to form into STL. The original Circuit dataset contains 120K NL-STL pairs, while we find including 8K examples into our dataset is informative enough to cover the whole corpus richness of Circuit dataset.

Hence, in this work a dataset with in total about 28K lifted NL-STL pairs are created. Appendix~\ref{sec:appendixE1} shows the statistics of this lifted NL-STL dataset.

\begin{figure}[t]
  \centering
  \includegraphics[width=0.9\linewidth]{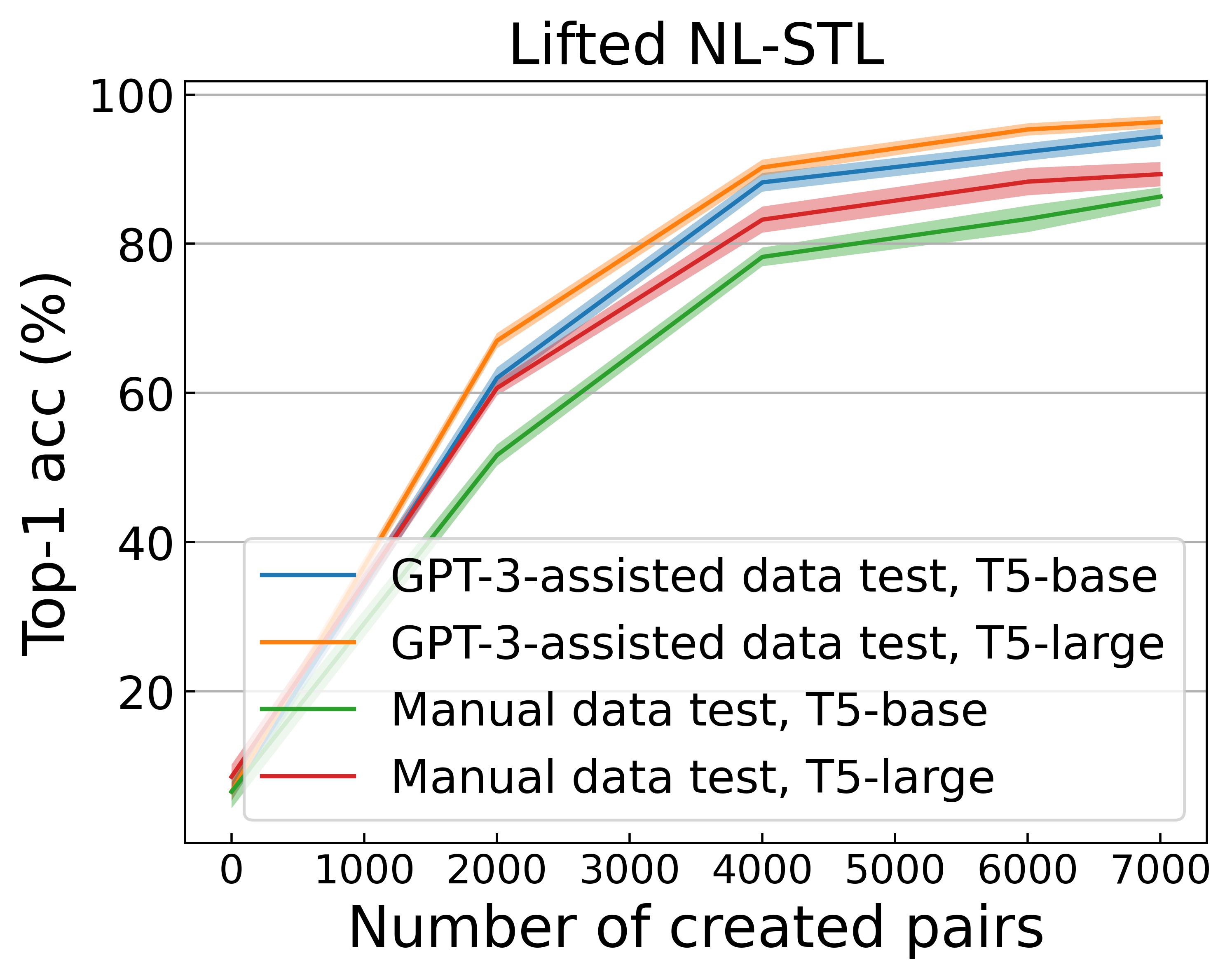}
   \caption{
Testing accuracy vs. Number of created NL-STL pairs. The data collected from Navigation and Circuit work are all used during training. The GPT-3-assisted data refers to the data generated with the help of GPT-3, and the Manual data refers to the instructions collected from volunteers. The figure shows the necessity of the created data.}
   \label{fig:fourfigure}
\end{figure}

\subsection{Model Finetuning}
\label{sec:model finetuning}
We mainly apply the T5 model \cite{raffel2020exploring} to serve as the base LLM for finetuning. To study whether model sizes will influence the performance, T5-base (220M) and T5-large (770M) are both finetuned on the same data. The setting is illustrated in Appendix~\ref{sec:appendix-details-of-implement}.

\section{Experimental Results}
\label{sec:experimental results}
Though the corrected data from Navigation and Circuit studies already provide multiple examples, these datasets only cover limited conditions and lack generalization. To show the necessity of the newly created data, the T5 models are finetuned with varied number of created NL-STL pairs, as shown in Figure~\ref{fig:fourfigure}. During training, all the data collected from Navigation and Circuit studies are used and the number of created data are varied among different models. The trained models are then tested with either the created data (referred as GPT-3-assisted data test) or the NL instructions collected from volunteers (referred as Manual data test). Since minor difference in STLs can cause severe difference in the real meanings, we apply the binary accuracy as the metric, i.e., 100\% right or not.  We find that the Top-1 testing accuracy increases greatly increasing the number of created pairs, with the highest accuracy 97.52\% and 90.12\% of GPT-3 assisted data and Manual data testing, respectively. 

Table~\ref{tab:result_table1} presents the experimental results with the targeted STL of different formats as discussed in Section~\ref{sec:STL formats}. We find that using the in-order format plus replacing operators with words will largely improve the performance. In-order format is more consistent with natural sentence expressions and lowers the difficulty for finetuning an LLM. This result is different from former conclusions when training Seq2Seq model for NL to STL/LTL tasks, where the pre-order format is better because it naturally avoids the issue of parentheses matching \cite{wang2021learning}.

\begin{table}
\centering
\begin{tabular}{ccc}
\toprule
\textbf{ } & \textbf{T5-base} & \textbf{T5-large}\\
\midrule
P.O./word & $70.00\pm1.42\%$ & $73.10\pm1.05\%$ \\
I.O./word & $\textbf{96.43}\pm\textbf{0.72}\%$ & $\textbf{97.52}\pm\textbf{0.65}\%$ \\
P.O./opera. & $72.35\pm1.54\%$ & $71.95\pm1.23\%$\\
I.O./opera. & $89.94\pm0.89\%$ & $88.17\pm1.02\%$\\
\bottomrule
\end{tabular}
\caption{Accuracy of GPT-3-assisted data testing for training data with different expression formats. P.O. and I.O. represent Pre-order and In-order, respectively.}
\label{tab:result_table1}
\end{table}

\begin{table*}
\centering
\begin{tabular}{cccc}
\toprule
\textbf{ } & \textbf{Circuit} & \textbf{Navigation} & \textbf{Office email}\\
\midrule
GPT-4 end-to-end (ad-hoc) & $62\%$ & $87\%$ & $84\%$ \\
GPT-3 end-to-end & $38.25\pm6.51\%$ & $50.51\pm5.08\%$ & $58.73\pm4.86\%$ \\
T5-large + GPT-3 AP detect & $\textbf{95.13}\pm\textbf{1.42}\%$ & $\textbf{95.03}\pm\textbf{1.20}\%$ & $\textbf{96.73}\pm\textbf{1.03}\%$ \\
T5-base + GPT-3 AP detect & $94.61\pm0.74\%$ & $94.73\pm1.02\%$ & $96.08\pm0.97\%$\\
\bottomrule
\end{tabular}
\caption{Testing accuracy of full NL-to-STL task for each grounding model. The testing domains are: Circuit \cite{he2022deepstl}, Navigation \cite{wang2021learning}, Office email \cite{aaai2023fc}.}
\label{tab:result_table2}
\end{table*}

\begin{table*}
\centering
\begin{tabular}{cccc}
\toprule
\textbf{ } & \textbf{Circuit} & \textbf{Navigation} & \textbf{Office email}\\
\midrule
GPT-3 AP detect accuracy & $98.84\pm0.41\%$ & $99.03\pm0.53\%$ & $100.00\pm0.00\%$\\
\bottomrule
\end{tabular}
\caption{Testing accuracy of recognizing APs with GPT-3 for each domain.}
\label{tab:result_table3}
\end{table*}

\section{Ablation Studies}
\textbf{Human Annotation}\quad To reveal the significance of human annotation, we train the model with the same amount of raw pairs created by GPT-3 and test them on corrected data. The results are shown in Appendix~\ref{sec:significance of human annotation}. We find that annotated dataset can improve the testing accuracy by around 10\%.
\\\\
\textbf{Framework2}\quad To reveal the significance of the data generated by Framework2, we train the model with either the same amount of data from pure Framework1 or the data combining two frameworks. Utilizing both frameworks improves the accuracy by around 2\% (Appendix~\ref{sec:significance of Framework2}).
\\\\
\textbf{Model Capacity of T5}\quad To reveal the significance of T5's superior model capacity, we train a Seq2Seq model on the same lifted NL-STL dataset for comparison, as shown in Appendix~\ref{sec:appendixmodelcapacity}. Finetuning on T5 model improves the accuracy by around 14.5\% compared to the Seq2Seq model.

\section{Application}\label{application}
Right now we have finetuned T5 model to convert lifted NL into lifted STL. For the real applications, we need one model to convert from full NL to full STL, in which the format of APs should be regularized. To reach this destination, we will display two methods in the following discussion, and compare them with other state-of-the-art models. We test on five datasets across domains Circuit \cite{he2022deepstl}, Navigation \cite{wang2021learning}, Office email \cite{aaai2023fc}, GLTL \cite{gopalan2018sequence, tellex2020robots}, and CW \cite{squire2015grounding}. The examples of full NL-STL pairs in each domain are shown in Appendix~\ref{sec:appendixD}, and the statistics of our synthesized dataset and each collected dataset are shown in Appendix~\ref{sec:appendixE2}. Note that some lifted NL-STL pairs in Circuit and Navigation datasets have been used during training the lifted model, while all the full NL-STL pairs have not been seen. All the data in other three domains are independent of the finetuning in lifted models. Our model achieves higher accuracy on full NL-STL transformation with much less training data across all these domains.

\subsection{Lifted Model + GPT-3 AP Recognition}
\label{subsection:AP recognition}
In the real applications, we have to formulate how the APs are presented in STL (like 'verb\_noun') so that the specified APs can directly connect with controllers. As shown in Appendix~\ref{sec:appendix-ap-recognition}, we directly utilize GPT-3 to recognize APs in the sentence and hide them as \emph{"prop\_i"}. Then the lifted model will predict the targeted lifted STL and the hidden APs will be swapped into formatted form to generate the full STL.

Table~\ref{tab:result_table2} displays the performance accuracy of this method. We test on three distinct domains and compare with the GPT-3 end-to-end method, i.e., using GPT-3 to directly transform NL into STL. The GPT-3 end-to-end method is proposed by \citet{aaai2023fc} recently, aiming to generalize into all various domains. However, in the NL to STL/LTL task, finetuning on a much smaller LLM like T5 is still greatly better than direct few-shot learning on state-of-the-art LLM like GPT-3 and GPT-4. Due to the limitation of GPT-4 access, we did an ad-hoc test on ChatGPT Plus with 100 samples in each domain. The experimental results show that combining finetuned lifted model with AP recognition using GPT-3 can lead to a full task accuracy over 95\% across all three tested domains. Table~\ref{tab:result_table3} displays the performance of detecting APs with GPT-3. Compared to the direct NL to STL task, AP detection task is much easier to GPT-3. Hence, dividing the whole task into AP recognition and semantic parsing are more data-efficient and flexible than pure end-to-end method.

To further test model performance under varied sentence complexity, we plot the testing accuracy vs. the number of APs in Appendix~\ref{sec:appendix-acc-vs-APnum}. As the number of APs in each lifted STL increases, the accuracy of GPT-3 few-shot learning decreases, while the finetuned T5-large model still performs well.

\subsection{Transfer Learning}
\label{subsection:transfer learning}
\begin{figure*}[t]
  \centering
  \includegraphics[width=0.95\linewidth]{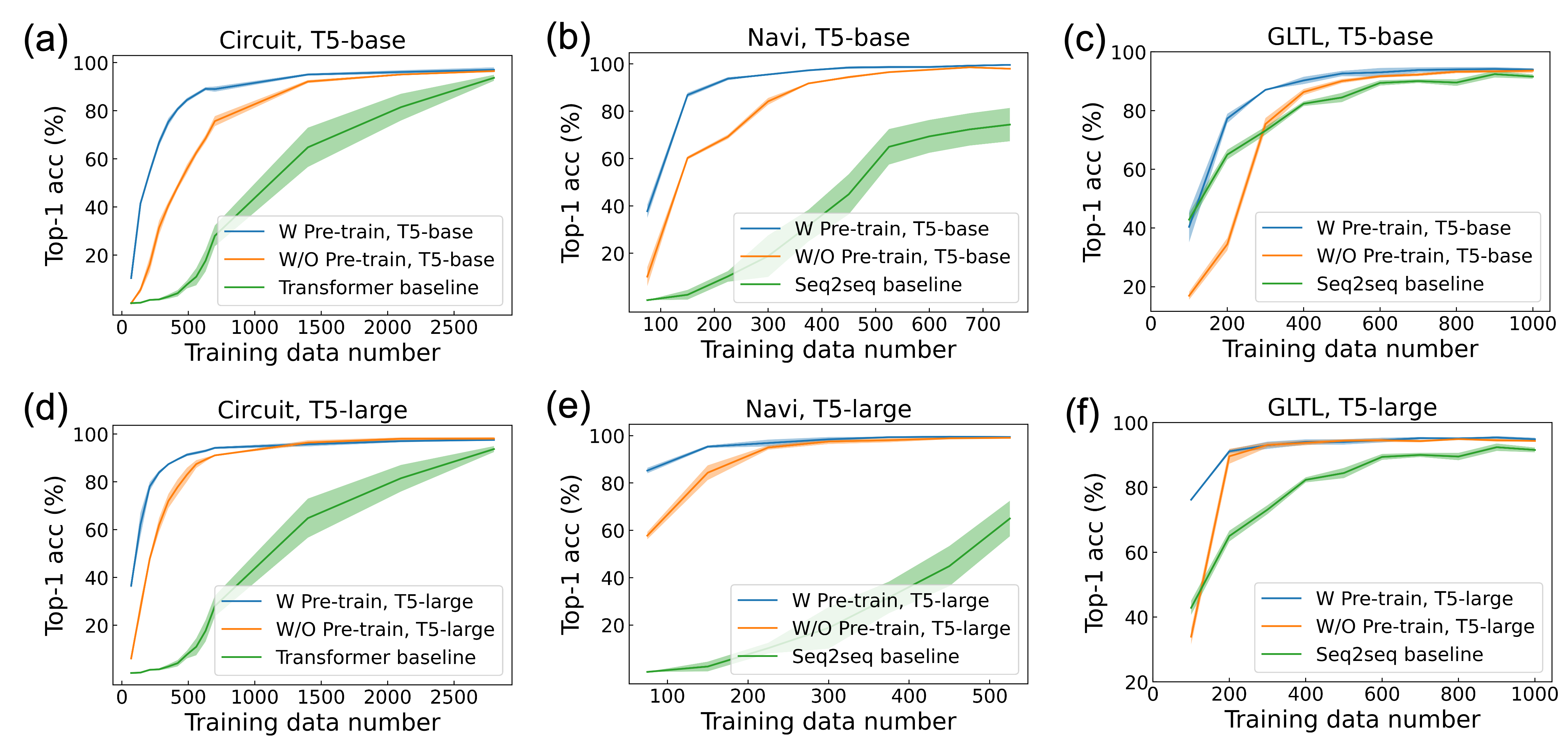}
   \caption{
Experimental results for end-to-end transfer learning on Circuit, Navigation, and GLTL datasets. Here the training data number means how many full NL-STL pairs are used during transfer learning or Seq2Seq training. The blue curve represents the accuracy where T5 model first pre-trained on 28K lifted NL-STL pairs, and then finetuned on full NL-STL examples in that domain. The orange curve represents the condition when T5 model is not pre-trained by lifted NL-STL pairs, but directly finetuned based on initial released weights.}
   \label{fig:sevenfigure}
\end{figure*}

\begin{figure*}[t]
  \centering
  \includegraphics[width=0.7\linewidth]{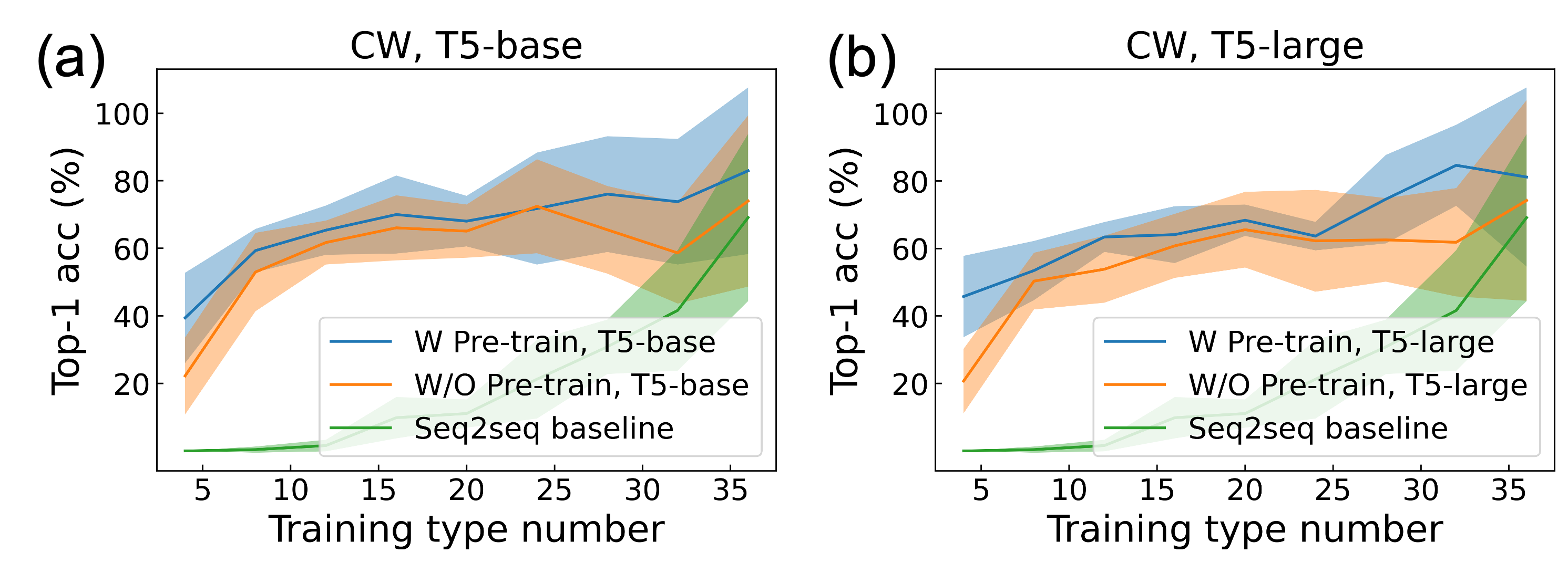}
   \caption{
Experimental results for end-to-end transfer learning on CW datasets. This experiment is to compare generalizability of our method with the original state-of-the-art Seq2Seq method.}
   \label{fig:eightfigure}
\end{figure*}

On the condition that we know how the users define the representation of APs, the aforementioned method is suitable to predict the full STL. On the other hand, there is also the condition that we cannot acquire the specific hard-coded rules to formulate AP representation, but only full NL-STL pairs. In these cases, the direct further finetuning may help. In other words, the lifted model has learnt to parse the semantic logical relations, and the further transfer learning is to learn how the APs are regulated in this specific dataset. This direct end-to-end transfer learning serves as the second way for ground applications.

To show that our method is generalizable and data-efficient, we compare our methods to the original Seq2Seq methods implemented in each dataset. Specifically, in the Circuit dataset the authors train the model from the ground using Transformer architecture \cite{vaswani2017attention}, and in GLTL and CW datasets the authors implement recurrent neural network (RNN) encoder-decoder framework with Gated Recurrent Unit (GRU) as the core RNN cell. As the work on Navigation dataset uses the final task completion rate as the criterion not the direct LTL accuracy, the LTL prediction accuracy is inherently low. For a fair comparison in Navigation dataset, we implements the same Seq2Seq framework as that in GLTL and CW datasets.

The experimental results are shown in Figure~\ref{fig:sevenfigure}. Compared to the original Seq2Seq model proposed in each dataset, transfer learning with LLM is much more efficient, and the pre-training on lifted NL-STL pairs also displays a great saving on training data requirements. We also find that T5-large model performs better than T5-base model. In all the three domains, the T5-large model with lifted NL-STL pre-training can achieve an accuracy near 95\% with only 200 to 500 full NL-STL examples. This amount of example requirement is one magnitude less than the Seq2Seq baselines.

The CW dataset is somewhat unique since it only has 36 different LTLs, meaning there are on average 50 different NLs corresponding to the same LTL. The study in \citet{gopalan2018sequence} applies this dataset to test the generalizability of the models within the domain. They use some types of LTLs as the training examples for transfer learning, and the left types of LTLs as the testing set. This is to test whether the model can predict the LTLs that it has not seen during the training. We also carry out this experiment and compare with the method in the original paper. As shown in Figure~\ref{fig:eightfigure}, the LLM with finetuning is apparently better than the original baseline.

\section{Limitation}
In spoken language, coreference is quite common, such as \emph{"pick up the apple and then bring it to me"}. Here \emph{"apple"} and \emph{"it"} refer to the same object. In the five datasets we collected and tested, the coreference problem is not severe since most NL do not have pronouns. For further work, the NER models specialized in resolving coreferences and converting them into normal APs are needed for more unbounded input sentences.

During the synthesis of varied STLs, here we use direct algorithm-based method to generate STL binary trees. To make the semantic meanings of STLs closer to human spoken language, we add the extra NL-STL loop via GPT-3. However, another intuitive way is to fit the probable distribution of operators to be close to human spoken language. For instance, the probability of two continued 'negation' operators is nearly zero. In this work we only set some hard rules to specify the STL synthesis. Further work can focus on the fitting of operator distributions and apply it into STL generation.

The evaluation metric here is pure binary accuracy (fully correct or not). Actually, it is quite difficult to judge the similarity or distance of two TLs. Simply calculating token matching or computing truth values both own drawbacks. A more effective metric is needed.

The output of LLMs may sometimes generate incorrect TLs. We build up rule-based methods to check syntactic correctness and correct errors like parentheses matching. Further work can be added to improve output correctness by modifying training procedures and loss functions.

\section{Conclusion}
We propose a framework to achieve NL-to-TL transformation with the assistance of LLM, from aspects of both data generation and model training. One dataset with about 28K lifted NL-TL pairs is then constructed by which the T5 model is finetuned. Two approaches are implemented to utilize the trained model into full NL-to-TL translation. Experimental results on five varied domains display much better accuracy and generalizablity compared to original methods. The created dataset can be used to train future NL-to-TL models and serve as the benchmark. 
The proposed framework to finetune LLMs with lifted NL-TL pairs makes it possible for generalizable NL-to-TL translation without the constraints of domains and input instruction structures. 

Although our framework is built on GPT-3, the rapid progression of LLMs can promote our framework. The strong semantic parsing ability of newly released GPT-4 will mitigate the burden of human annotations in our method. We find that GPT-4 generated STLs/NLs are closer to the correct answers, compared to GPT-3 generated STLs/NLs. As future work, we believe the model can be improved with larger dataset containing more diversified corpus with GPT-4 as the base model.

\section*{Acknowledgements}

We thank the help from the volunteers for contributing the natural language instructions.

This work was supported by ONR under Award N00014-22-1-2478 and MIT-IBM Watson AI Lab. However, this article solely reflects the opinions and conclusions of its authors. The authors would also like to thank Lifu Huang, Zhiyang Xu, and Yue Meng for the early-stage exploration and discussion of the work.

\bibliography{anthology,custom}
\bibliographystyle{acl_natbib}

\newpage

\appendix
\onecolumn
\section{STL illustration}
\begin{table}[h]
\centering
\begin{tabular}{p{0.1\linewidth} | p{0.75\linewidth}}
\hline
$\textbf{F}_{[a,b]}\phi$ & True at time $t$ if there exists a time in the interval $[t+a, t+b]$ where $\phi$ is true.\\
\hline
$\phi\textbf{U}_{[a,b]}\varphi$ & True at time $t$ if $\varphi$ is true for some time $t^{'}$ in the interval $[t+a, t+b]$, and for all times between $t$ and $t^{'}$, the formula $\phi$ holds.\\
\hline
$\textbf{G}_{[a,b]}\phi$ & True at time $t$ if for all times in the interval $[t+a, t+b]$, the formula $\phi$ holds.\\
\hline
\end{tabular}
\caption{STL illustration}
\label{tab:STL}
\end{table}

\newpage
\section{Full algorithm to synthesize multiple STLs}
\label{sec:appendixalgo}
This is the full algorithm to synthesize multiple varied STLs. The blue-colored words are the example output in each step. All the operators are classified into the operator with only one subtree, or the operator with two subtrees. A random ordered $prop$ list is generated with the length less than the upper limit. Then this full list is split into some sub\_lists. For each sub\_list, operators are randomly appended in the left side until each $prop$ occupy one position in the binary tree. Then these modified sub\_lists are assembled back into the full STL by appending operators with two subtrees. The STL generated in this way are syntactically correct, but may own some flaws in semantic meanings. Some rules are pre-set to avoid the unreasonable conditions, e.g., two negation operation should not appear continually.

\begin{algorithm}
\caption{Full algorithm for STL synthesis}\label{alg:full}
\begin{algorithmic}[1]
 \renewcommand{\algorithmicrequire}{\textbf{Input:}}
  \renewcommand{\algorithmicensure}{\textbf{Output:}}
\Require 
{\\ Maximum number of APs $N$}
\Ensure {\\ Synthesized pre-order STL}
\\
\\ $two\_subtree$ = [$\land$, $\lor$, $\Rightarrow$, $\Leftrightarrow$, $\textbf{U}$, $\textbf{U}_{[a,b]}$],
\\ $one\_subtree$ = [$\neg$, $\textbf{F}$, $\textbf{G}$, $\textbf{F}_{[a,b]}$, $\textbf{G}_{[a,b]}$]
\\ AP\_num = random.randint(1, $N$) \textcolor{blue}{\Comment{e.g., 3}}
\\ prop\_list $\gets$ Random\_ordered Prop list with length AP\_num
\\ sub\_lists $\gets$ randomly divide prop\_list \textcolor{blue}{\Comment{e.g., [prop\_3, prop\_1], [prop\_2]}}
\\ 
\\ \textit{\textbf{Creating sub-STLs}} :
\For{each sub\_list}
    \State num\_open\_subtree = len(sub\_list)
    \While{num\_open\_subtree > 1}
        \State operation $\gets$ randomly choose item in $two\_subtree+one\_subtree$
        \If{operation in $two\_subtree$}
            \State num\_open\_subtree -= 1
        \EndIf
        
        \If{operation in [$\textbf{U}_{[a,b]}$, $\textbf{F}_{[a,b]}$, $\textbf{G}_{[a,b]}$]}
            \State $a, b$ $\gets$ sampling random integers or denoting as $infinity$      
        \EndIf
        \State sub\_list.insert(0, operation)
    \EndWhile
    \State save sub\_list as sub\_STL \textcolor{blue}{\Comment{e.g., [$\Leftrightarrow$, $\neg$, prop\_3, prop\_1], [$\textbf{G}$, prop\_2]}}
\EndFor
\\
\\ \textit{\textbf{Assembling sub-STLs}} :
\\ Assembling sub\_STLs into pre-order STL by appending random $two\_subtree$ operations
\textcolor{blue}{\Comment{e.g., [$\textbf{U}_{[10,30]}$, $\Leftrightarrow$, $\neg$, prop\_3, prop\_1, $\textbf{G}$, prop\_2]}}
\end{algorithmic}
\end{algorithm}

\section{Examples of prompt input to GPT-3}
These are the example prompts for GPT-3 to convert between NL and STL, or detect the spans of Atomic Proportions.
\newpage
\subsection{Prompt example from in-order STL to NL via GPT-3}
\label{sec:appendixA1}
\begin{figure*}[t]
  \centering
  \includegraphics[width=0.9\linewidth]{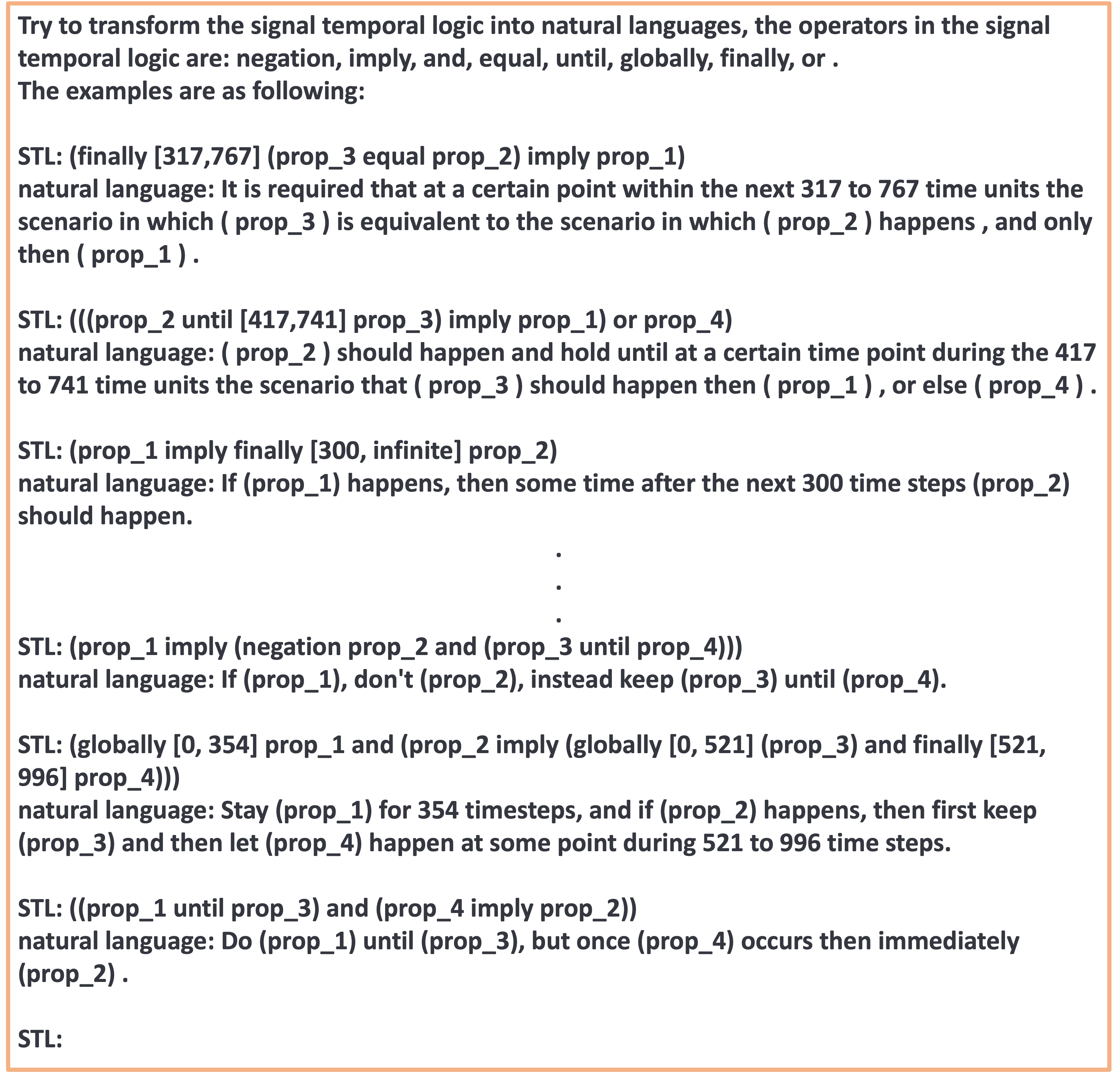}
   \caption{
Prompts for converting from synthesized STL to NL via GPT-3.}
   \label{fig:figureS1}
\end{figure*}
Figure~\ref{fig:figureS1} is a prompt example for GPT-3 to convert from STL to its corresponding NL. The input STL follows in-order expression with all the operators replaced by words with the same meanings. The prompt contains 20 NL-STL pairs, which are randomly picked up from 100 examples and are changed constantly during data creation. 

\newpage
\subsection{Prompt example from NL to pre-order STL via GPT-3}
\label{sec:appendixA2}
\begin{figure*}[t]
  \centering
  \includegraphics[width=0.9\linewidth]{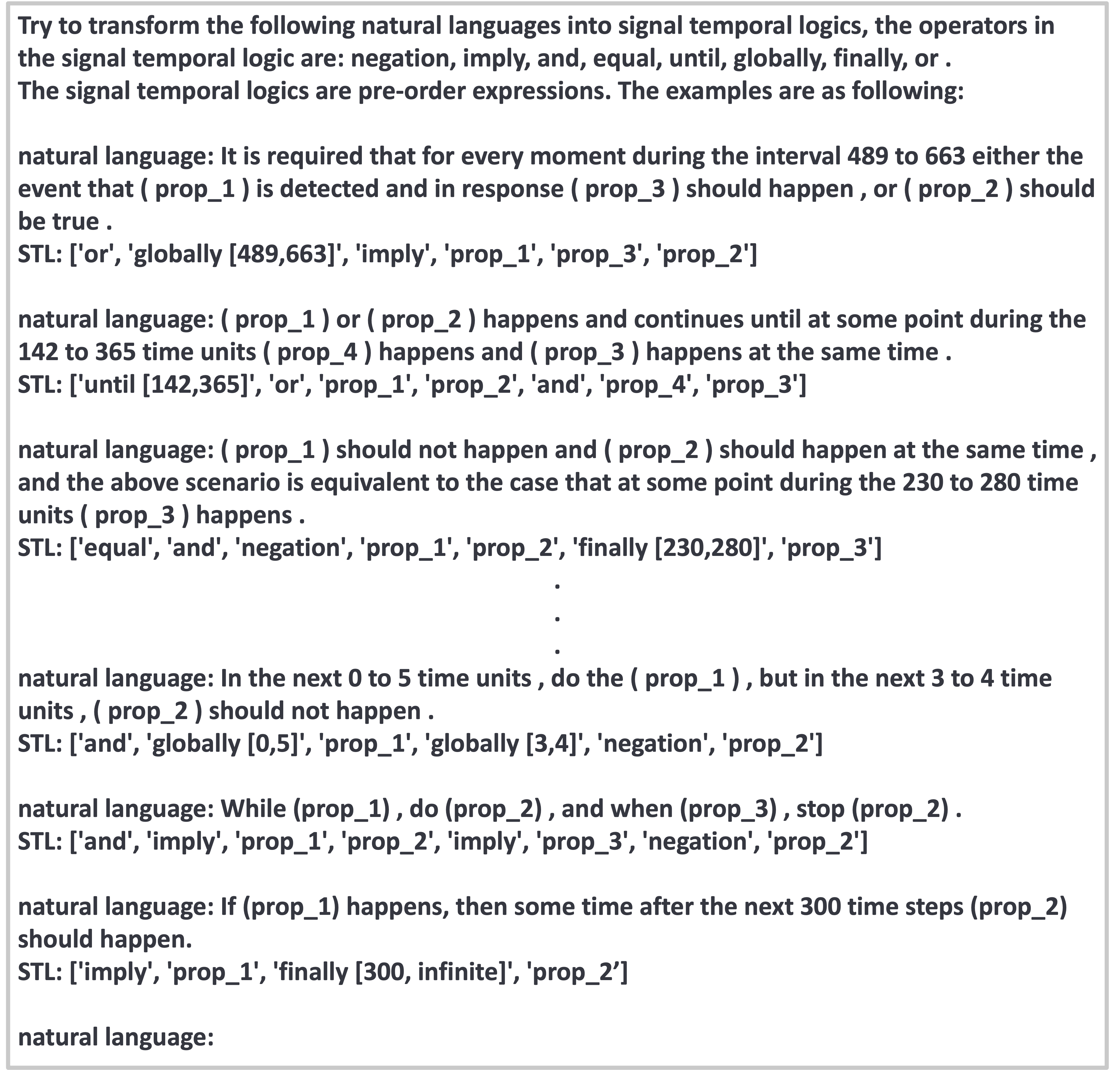}
   \caption{
Prompts for converting from NL to pre-order STL via GPT-3.}
   \label{fig:figureS2}
\end{figure*}
Figure~\ref{fig:figureS2} is a prompt example for GPT-3 to convert from NL to its corresponding STL. The output STL follows pre-order expression. We have tested that GPT-3 acts with close performance when STL follows either pre-order or in-order formats. 

\newpage
\subsection{Prompt example for AP recognition via GPT-3}
\label{sec:appendixA3}
\begin{figure*}[t]
  \centering
  \includegraphics[width=0.8\linewidth]{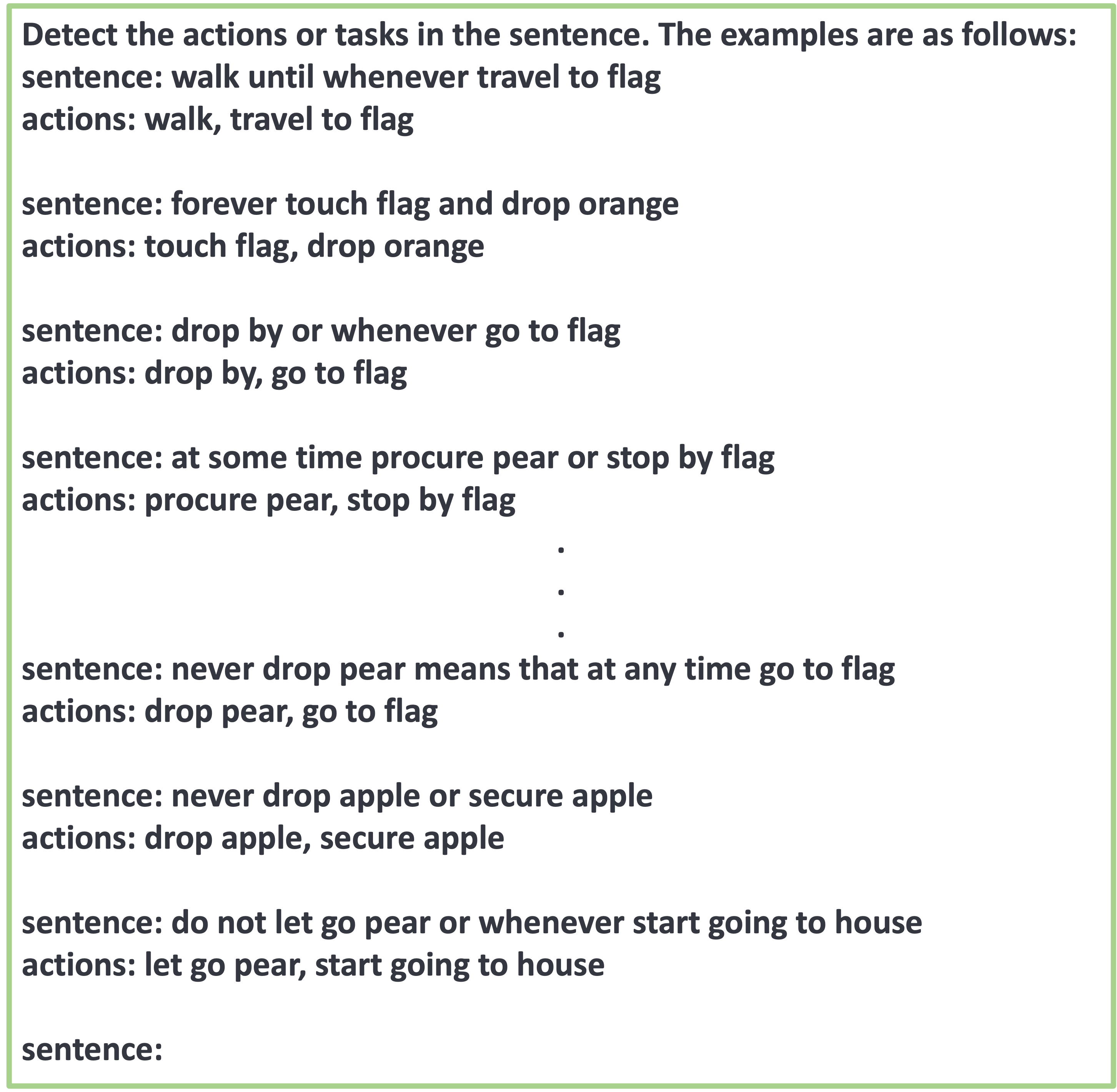}
   \caption{
Prompts for AP recognition via GPT-3.}
   \label{fig:figureS3}
\end{figure*}
Figure~\ref{fig:figureS3} is a prompt example for applying GPT-3 to detect APs in natural sentences. In this example, the specific domain is Navigation.

\newpage
\section{Number of NL-STL pairs in GPT-3 prompts}
\label{sec:appendix pair num vs acc}
\begin{figure*}[t]
  \centering
  \includegraphics[width=0.5\linewidth]{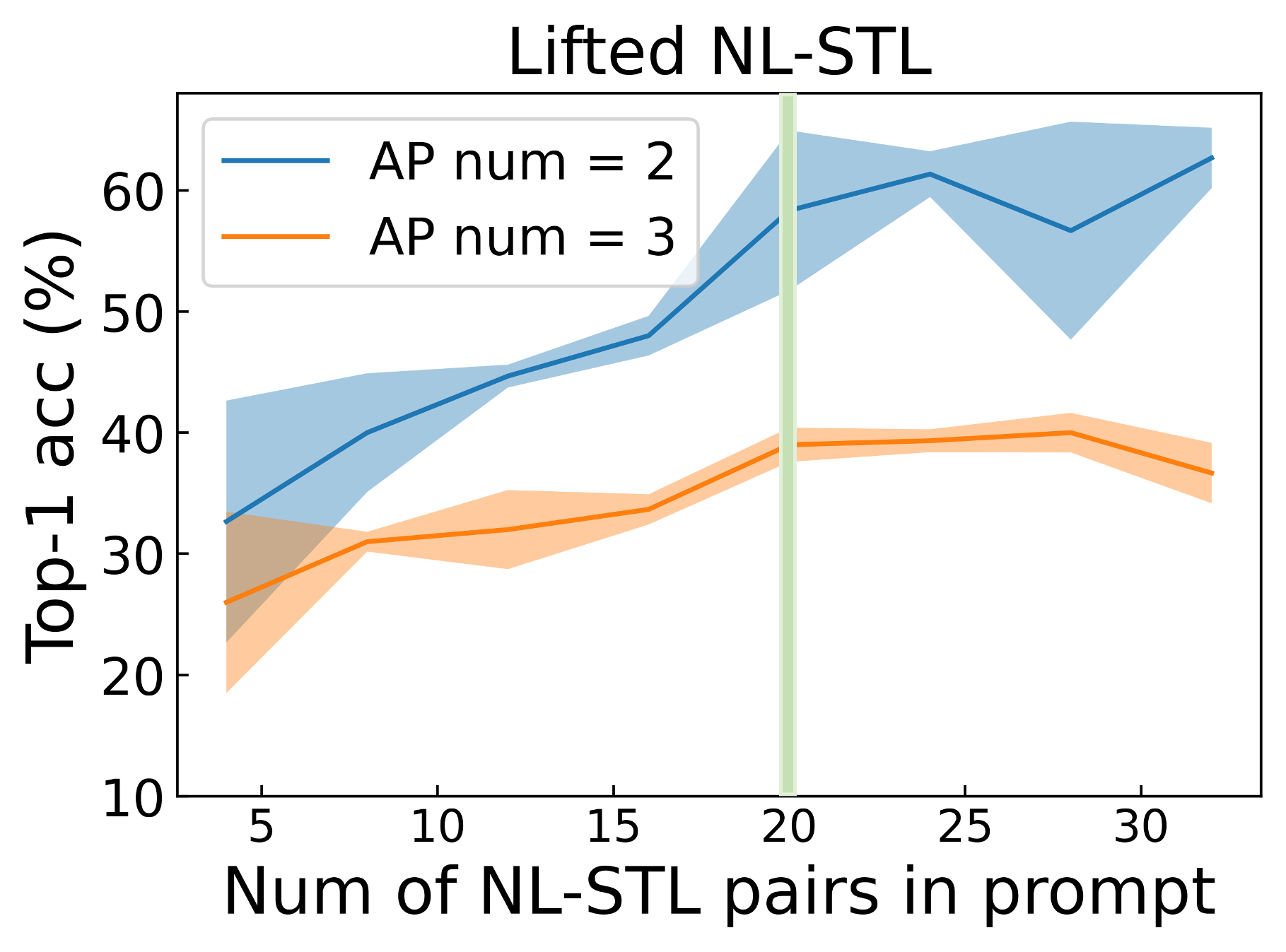}
   \caption{
Number of prompt pairs vs. GPT-3 performance.}
   \label{fig:prompt_pair_num_NL2STL}
\end{figure*}
As shown in Figure~\ref{fig:prompt_pair_num_NL2STL}, here we ask GPT-3 to transform from NL to STL and tune the number of NL-STL pairs in the prompt to detect the accuracy evolution. We test on the NL whose targeted STL have the number of APs to be two or three. We find that the prediction accuracy given by GPT-3 will arise with the number of example pairs and turn into a plateau when the number of example pairs increases to larger than 20. Here we choose the number of pairs to be 20 in the prompt.

\newpage
\section{Example annotations of lifted NL-STL pairs}
As shown in Table~\ref{tab:result_tableS1}.
\label{sec:appendixC}
\begin{table*}
\centering
\begin{tabular}{p{0.3\linewidth} | p{0.65\linewidth}}
\hline
\textbf{STL (pre-order+operator)} & ['<->', '->', 'prop\_2', 'prop\_3', 'F[55,273]', 'prop\_1']\\
\hline
\textbf{STL (in-order+word)} & ((prop\_2 imply prop\_3) equal finally[55,273] prop\_1)\\
\hline
\textbf{Raw natural sentence} & If (prop\_2) implies (prop\_3), then (prop\_1) will happen at some point during the next 55 to 273 time units .\\
\hline
\textbf{Annotated natural sentence} & If (prop\_2) implies (prop\_3), then (prop\_1) will happen at some point during the next 55 to 273 time units\textcolor{red}{, and vice versa .}\\\hline
{ } & { }\\
\hline
\textbf{STL (pre-order+operator)} & ['U[400,infinite]', '->', 'prop\_3', 'prop\_1', 'negation', 'prop\_2']\\
\hline
\textbf{STL (in-order+word)} & ((prop\_3 imply prop\_1) until[400,infinite] negation prop\_2)\\
\hline
\textbf{Raw natural sentence} & If (prop\_3), then do (prop\_1) and keep doing it until (prop\_2) happens, but this should never happen .\\
\hline
\textbf{Annotated natural sentence} & If (prop\_3), then do (prop\_1) and keep confirming to the above \textcolor{red}{state until (prop\_2) does not happens at some point after the 400 time units from now .}\\\hline
{ } & { }\\
\hline
\textbf{STL (pre-order+operator)} & ['<->', 'negation', 'prop\_1', 'U[279,438]', 'prop\_3', 'prop\_2']\\
\hline
\textbf{STL (in-order+word)} & (negation prop\_1 equal (prop\_3 until[279,438] prop\_2))\\
\hline
\textbf{Raw natural sentence} & The scenario in which ( prop\_1 ) happens is the same as the scenario in which ( prop\_3 ) happens and continues until at a certain time point during the 279 to 438 time units ( prop\_2 ) happens .\\
\hline
\textbf{Annotated natural sentence} & The scenario in which ( prop\_1 ) \textcolor{red}{does not happen} is the same as the scenario in which ( prop\_3 ) happens and continues until at a certain time point during the 279 to 438 time units ( prop\_2 ) happens .\\\hline
\end{tabular}
\caption{Example annotations from synthesized STLs to raw natural sentences, and further to annotated natural sentences.}
\label{tab:result_tableS1}
\end{table*}
\\

\section{Process of human annotation}
\label{sec:explanation for human annotation}
All the human annotators are researchers in the area of formal methods and robot planning with extensive knowledge in temporal logics. The annotators are all volunteers from the authors’ institute and collaborative institutes. Before the data annotation and collection, we have notified them that the data will be used to train the language model to transform from natural language instructions to temporal logics, and that both the annotated data and model will be open to the public. All the voluntary annotators have agreed to the statement to use their annotated data. As for the task instruction, the annotators participate in the guidance meeting and are provided with a guidance list on temporal logics and some example annotation pairs (the guidance list and examples of annotation are available on github page). Each annotator annotated 50 pairs initially, and sent their results to other randomly assigned annotators for cross-checking. Finally, the authors also checked all pairs to ensure the accuracy of the annotated data.
\\
\section{Example full NL-STL pairs of each specialized dataset}
As shown in Table~\ref{tab:result_tableS2}.
\label{sec:appendixD}
\begin{table*}
\centering
\begin{tabular}{p{0.1\linewidth} | p{0.05\linewidth} | p{0.7\linewidth}}
\hline
\textbf{Navigation} & \textbf{STL} & finally ( acquire\_v pear\_n ) and globally ( finally ( go\_to\_v waste\_basket\_n ) )\\
\hline
\textbf{ } & \textbf{NL} & when possible acquire pear and repeatedly go to waste basket .\\
\hline
\textbf{Navigation} & \textbf{STL} & finally ( got\_to\_v house\_n ) and finally ( go\_near\_v house\_n )\\
\hline
\textbf{ } & \textbf{NL} & at any time got to house and when possible go near house .\\
\hline
\textbf{Navigation} & \textbf{STL} & advance\_to\_v tree\_n imply finally ( get\_to\_v flag\_n )\\
\hline
\textbf{ } & \textbf{NL} & advance to tree means that when possible get to flag .\\
\hline
\textbf{Circuit} & \textbf{STL} & globally ( signal\_1\_n math equal 89.3 or ( signal\_2\_n more 42.4 and signal\_2\_n less 91.5 ) imply globally [0,34] ( finally [0,98] ( signal\_3\_n more equal 11.5 and signal\_3\_n less equal 23.4 ) ) )\\
\hline
\textbf{ } & \textbf{NL} & In the case the signal\_1\_n signal is 89.3 , or the signal\_2\_n signal is greater than 42.4 and below 91.5 , then for every time instant during the coming 34 time units , there needs to exist a certain time instant during the next 98 time units , at which the value of signal signal\_3\_n should be no less than 11.5 and less than or equal to 23.4 eventually .\\
\hline
\textbf{Circuit} & \textbf{STL} & finally ( signal\_1\_n less 92.6 and signal\_2\_n more equal 57.3 )\\
\hline
\textbf{ } & \textbf{NL} & At a certain time instant in the future before the end of the simulation signal\_1\_n is ultimately below 92.6 and signal\_2\_n will be ultimately at least 57.3 .\\
\hline
\textbf{Circuit} & \textbf{STL} & finally ( ( signal\_1\_n more equal 4.1 and signal\_1\_n less equal 59.0 ) or signal\_2\_n math equal 41.1 )\\
\hline
\textbf{ } & \textbf{NL} & There has to be a certain time instant in the future before the end of the simulation , at which the value of signal\_1\_n needs to be greater than or equal to 4.1 and less than or equal to 59.0 eventually , or signal\_2\_n finally keeps equal to 41.1 .\\
\hline
\textbf{GLTL} & \textbf{STL} & finally ( ( red\_room or blue\_room ) and finally green\_room )\\
\hline
\textbf{ } & \textbf{NL} & enter the blue or orange room and proceed until the green room .\\
\hline
\textbf{GLTL} & \textbf{STL} & ( finally ( blue\_room ) and globally ( negation green\_room ) )\\
\hline
\textbf{ } & \textbf{NL} & move to the blue room without entering a lime room .\\
\hline
\textbf{GLTL} & \textbf{STL} & ( finally ( yellow\_room ) and globally ( negation blue\_room ) )\\
\hline
\textbf{ } & \textbf{NL} & only go through rooms that are not purple to get to the yellow room .\\
\hline
\textbf{CW} & \textbf{STL} & finally ( blue\_room and finally green\_room )\\
\hline
\textbf{ } & \textbf{NL} & please go to the green room through the blue room .\\
\hline
\textbf{CW} & \textbf{STL} & finally red\_room\\
\hline
\textbf{ } & \textbf{NL} & i want you to go into the red room .\\
\hline
\textbf{CW} & \textbf{STL} & finally ( ( red\_room or yellow\_room ) and finally green\_room )\\
\hline
\textbf{ } & \textbf{NL} & go thru the yellow or red box to get to the green box .\\
\hline
\textbf{Office email} & \textbf{STL} & globally ( ( ( a new incident is created in Eventribe ) and ( a response is created in Trello ) ) imply ( creating an object in Gmail ) )\\
\hline
\textbf{ } & \textbf{NL} & When the transition action that a new incident is created in Eventribe does not get observed , and a response is created in Trello , then the following condition is true : promptly creating an object in Gmail .\\
\hline
\textbf{Office email} & \textbf{STL} & ( ( sync Microsoft Teams data ) until finally ( sending me an SAP and Salesforce ) )\\
\hline
\textbf{ } & \textbf{NL} & sync Microsoft Teams data until when possible sending me an SAP and Salesforce .\\
\hline
\textbf{Office email} & \textbf{STL} & globally ( ( ( a new lead is added in Marketo ) and ( creating a new Marketo card ) ) imply ( a new lead is added in Microsoft Teams ) )\\
\hline
\textbf{ } & \textbf{NL} & On condition that a new lead is added in Marketo and creating a new Marketo card , then the event that a new lead is added in Microsoft Teams needs to occur at the same time instant .\\
\hline
\end{tabular}
\caption{Examples of full NL-STL pairs in each specialized domain.}
\label{tab:result_tableS2}
\end{table*}

\newpage
\section{Ablation Studies}
\subsection{Significance of Human Annotation}
\label{sec:significance of human annotation}
\begin{figure}[t]
  \centering
  \includegraphics[width=0.5\linewidth]{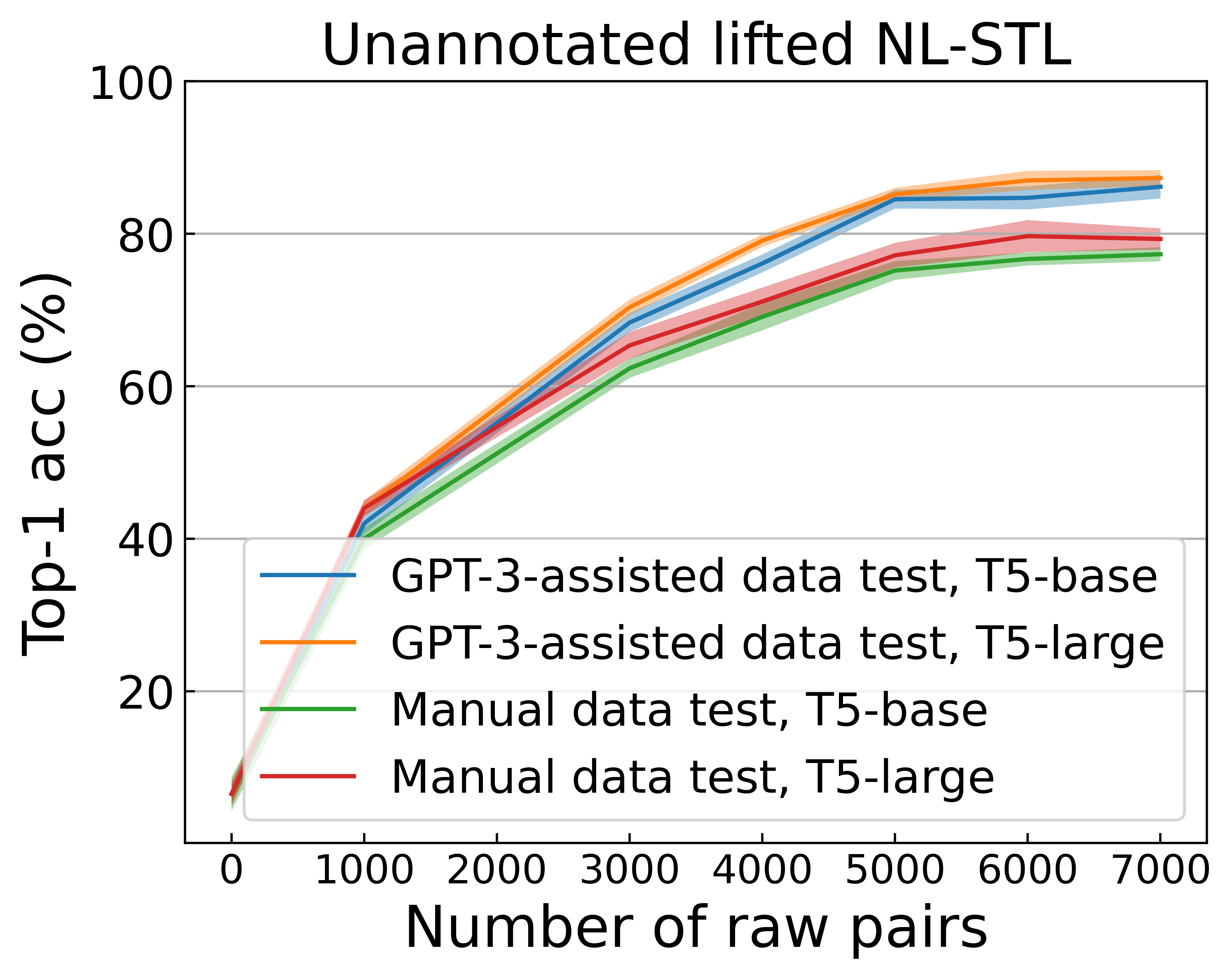}
   \caption{
Testing accuracy vs. Number of raw NL-STL pairs. The data collected and re-annotated from Navigation and Circuit work are all used during training. The GPT-3-assisted data refers to the data generated with the help of GPT-3, and the Manual data refers to the instructions collected from volunteers.}
   \label{fig:pure raw training data}
\end{figure}

This part is to demonstrate the significance of human annotation for the GPT-3 synthesized data. Figure~\ref{fig:pure raw training data} shows the model accuracy under varied number of training raw pairs. The great thing is that the T5-large model can still achieve a highest testing accuracy of 87.3\% and 79.4\% on the GPT-3-assisted data and Manual data test, even only using the raw data synthesized from GPT-3. However, compared to the results in Figure~\ref{fig:fourfigure}, models trained on annotated data achieves accuracy about 10\% higher than models trained on raw data.
\\
\subsection{Significance of Framework2}
\label{sec:significance of Framework2}

\begin{table*}[h]
\centering
\begin{tabular}{ccccc}
\toprule
\multicolumn{1}{c}{} & \multicolumn{2}{c}{\textbf{3K dataset}} & \multicolumn{2}{c}{\textbf{4.5K dataset}} \\
\cmidrule(rl){2-3} \cmidrule(rl){4-5}
\textbf{Domain} & {1.5K F1 + 1.5K F2} & {3K F1} & {3K F1 + 1.5K F2} & {4.5K F1} \\
\midrule
Raw data & $\textbf{78.85}\pm\textbf{1.04}\%$ & $75.79\pm0.98\%$ & $\textbf{80.48}\pm\textbf{0.71}\%$ & $79.04\pm0.64\%$ \\
Annotated data & $\textbf{80.57}\pm\textbf{0.86}\%$ & $79.76\pm0.88\%$ & $\textbf{88.32}\pm\textbf{0.84}\%$ & $86.51\pm0.77\%$ \\
\bottomrule
\end{tabular}
\caption{Testing accuracy of the models with different training datasets. The training data are either raw or annotated, pure from Framework1 (F1) or combining with Framework2 (F2). The experimental results show that the annotated dataset can apparently improve the performance of the model, and models combining the data generated by F1 and F2 outperform the models trained with the same amount of pure F1 data.}
\label{tab:significance of framework2}
\end{table*}

\newpage
\section{Model Capacity}
\label{sec:appendixmodelcapacity}
\begin{figure}[t]
  \centering
  \includegraphics[width=0.5\linewidth]{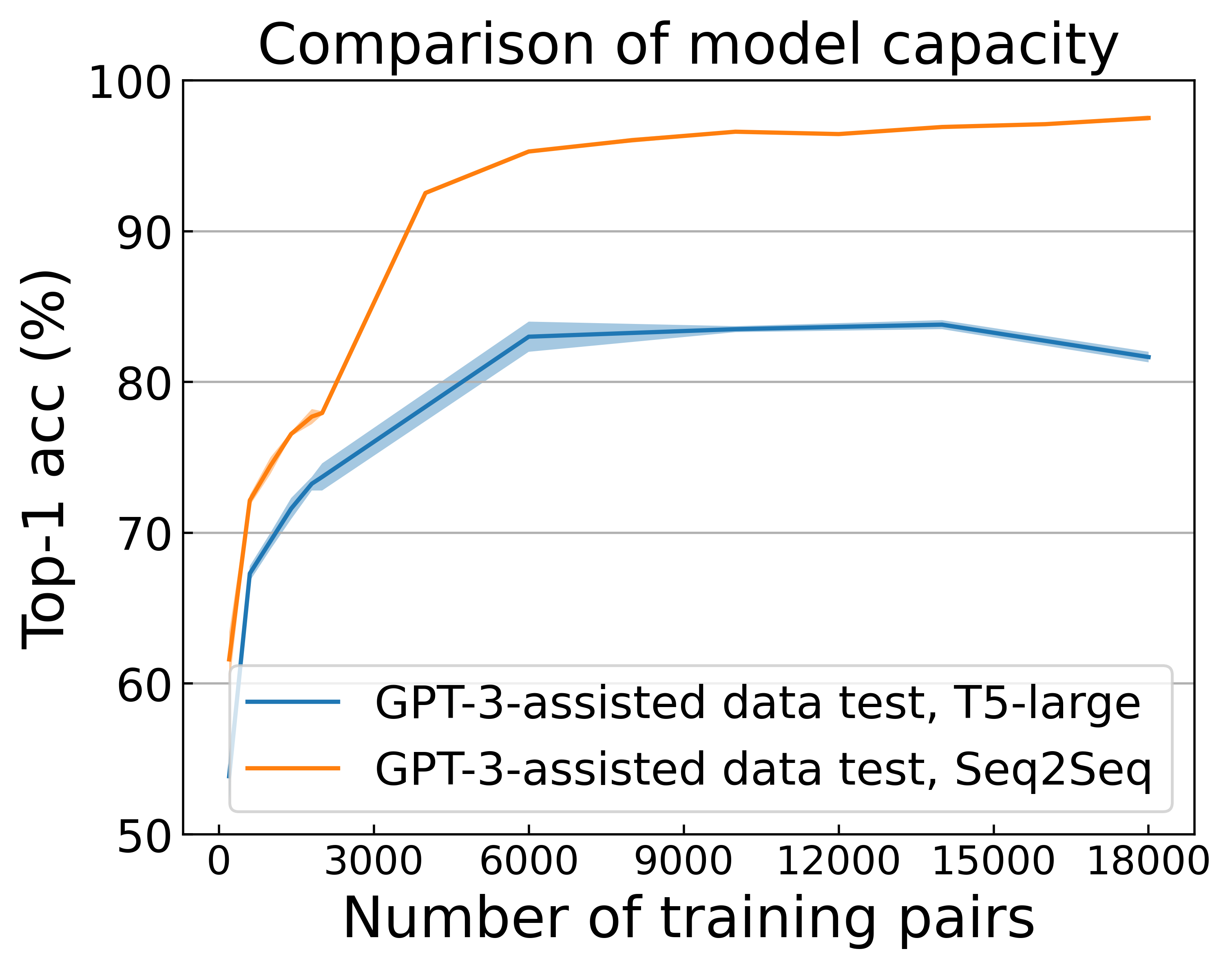}
   \caption{
Testing accuracy vs. Number of training lifted NL-STL pairs. Here we use T5-large and Seq2Seq models to train on the lifted data. We detect that the Seq2Seq model reaches a highest accuracy at 83\%, while T5-large model reaches a highest accuracy at 97.5\%.}
   \label{fig:model capacity figure}
\end{figure}

As shown in Figure~\ref{fig:model capacity figure}, T5-large performs much better than Seq2Seq model when training on the same lifted dataset. This reveals the significance to use LLM in this NL-to-TL task.

\newpage
\section{Datasets statistics}
\subsection{Statistics of lifted NL-STL dataset}
\label{sec:appendixE1}
\begin{table}[h]
\centering
\begin{tabular}{cccccc}
\toprule
\multicolumn{3}{c}{\textbf{\# APs per STL}} & \multicolumn{3}{c}{\textbf{\# Operators per STL}} \\
\cmidrule(rl){1-3} \cmidrule(rl){4-6}
avg. & median & max & avg. & median & max \\
\midrule
2.906 & 3 & 7 & 3.206 & 3 & 8 \\
\bottomrule
\end{tabular}
\caption{Lifted STL formula statistics: \# APs for each formula, \# STL operators for each formula.}
\label{tab:result_tableS3}
\end{table}

\begin{table}[h]
\centering
\begin{tabular}{cccccc}
\toprule
\multicolumn{1}{c}{} & \multicolumn{1}{c}{} & \multicolumn{4}{c}{\textbf{\# Words per Sent.}} \\
\cmidrule(rl){3-6}
\textbf{\# Sent.} & \textbf{\# Vocab} & {avg.} & {median} & {max} & {min} \\
\midrule
28466 & 2296 & 18.358 & 17 & 72 & 3 \\
\bottomrule
\end{tabular}
\caption{Lifted sentence statistics: \# unique sentences, \# unique words (vocab), \# words per sentence.}
\label{tab:result_tableS4}
\end{table}

\subsection{Statistics of corpus richness}
\label{sec:appendixE2}
\begin{table*}[h]
\centering
\begin{tabular}{cccc}
\toprule
\textbf{Domain} & \textbf{\# STL/ \#Sent.} & \textbf{\# Unique STL} & \textbf{\# Vocab.}\\
\midrule
Synthesized dataset & 15K & 14438 & 2121 \\
Circuit \cite{he2022deepstl} & 120K & 3653 &265 \\
Navigation \cite{wang2021learning} & 5K & 149 & 131 \\
GLTL \cite{gopalan2018sequence} & 11K & 193 & 193 \\
CW \cite{squire2015grounding} & 3.3K & 39 & 188 \\
Office email \cite{aaai2023fc} & 0.15K & 23 & 143 \\
\bottomrule
\end{tabular}
\caption{Statistics of STL formulas and NL sentences in our synthesized dataset and the collected dataset of each domain. \# STL/ \#Sent. reveals total number of samples. \# Unique STL counts the number of different STL formulas. \# Vocab. counts the vocabulary in each dataset. Compared to previously collected dataset, our synthesized dataset owns much larger number of unique STLs and vocabulary, revealing greater corpus richness.}
\label{tab:result_tableS5}
\end{table*}

\newpage
\section{Accuracy evolution with AP number}
\label{sec:appendix-acc-vs-APnum}
\begin{figure}[t]
  \centering
  \includegraphics[width=0.5\linewidth]{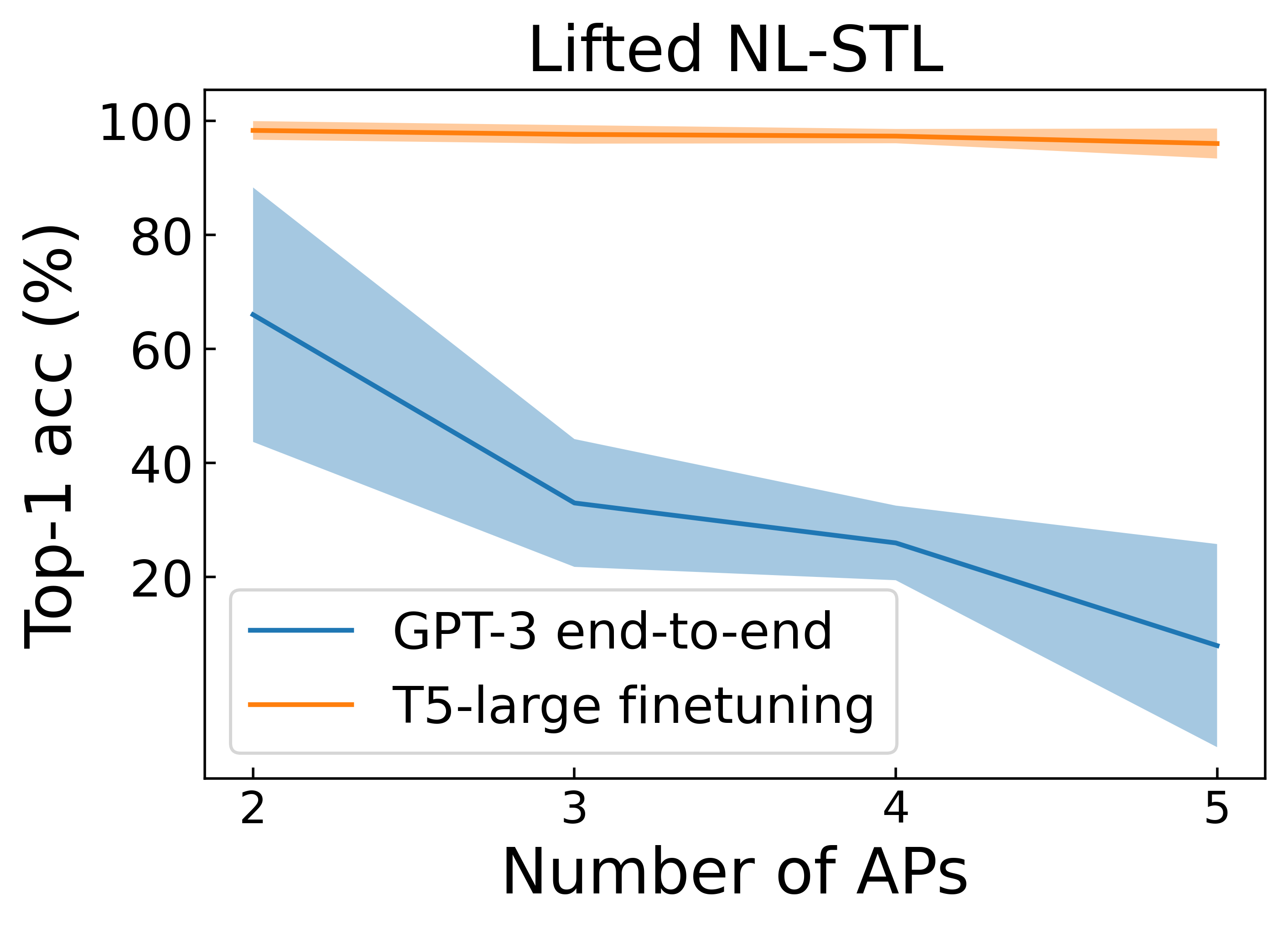}
   \caption{
Accuracy vs. number of APs in each lifted STL. We carry out the test with both finetuned lifted T5-model and GPT-3 end-to-end method.}
   \label{fig:figure_APnum_acc}
\end{figure}

This section is to illustrate that directly applying GPT-3 to predict STL from NL via few-shot learning largely decreases the accuracy when the sentence structure is complex. Here we hypothesize that sentence complexity is positively related to the number of APs. As shown in Figure~\ref{fig:figure_APnum_acc}, the prediction accuracy decreases rapidly with increasing AP number using GPT-3 end-to-end method. On the other hand, the method to finetune the T5-large using synthesized NL-STL pairs remains high accuracy across different AP numbers.
\\
\section{Details of implementation}
\label{sec:appendix-details-of-implement}
For all the finetuning experiments on both T5-base and T5-large models, we choose the learning rate as 2e-5, a batch size of 16, a weight decaying ratio as 0.01, and run 20 epochs for each setting. Experiments on average finish in 3 hours for T5-base, and 10 hours for T5-large, on a single Nvidia RTX 8000 GPU. Average results and standard deviations are typically acquired from 3 runs with seeds [1203, 309, 316], apart from the transfer learning in CW dataset where 10 runs are carried with seeds [1203, 309, 316, 34, 64, 128, 256, 512, 1234, 234]. For the finetuning on lifted models, the input dataset is split into training set (0.9) and testing set (0.1).

\newpage
\section{Full STL conversion by combining lifted model with AP recognition}
\label{sec:appendix-ap-recognition}
\begin{figure}[t]
  \centering
  \includegraphics[width=0.5\linewidth]{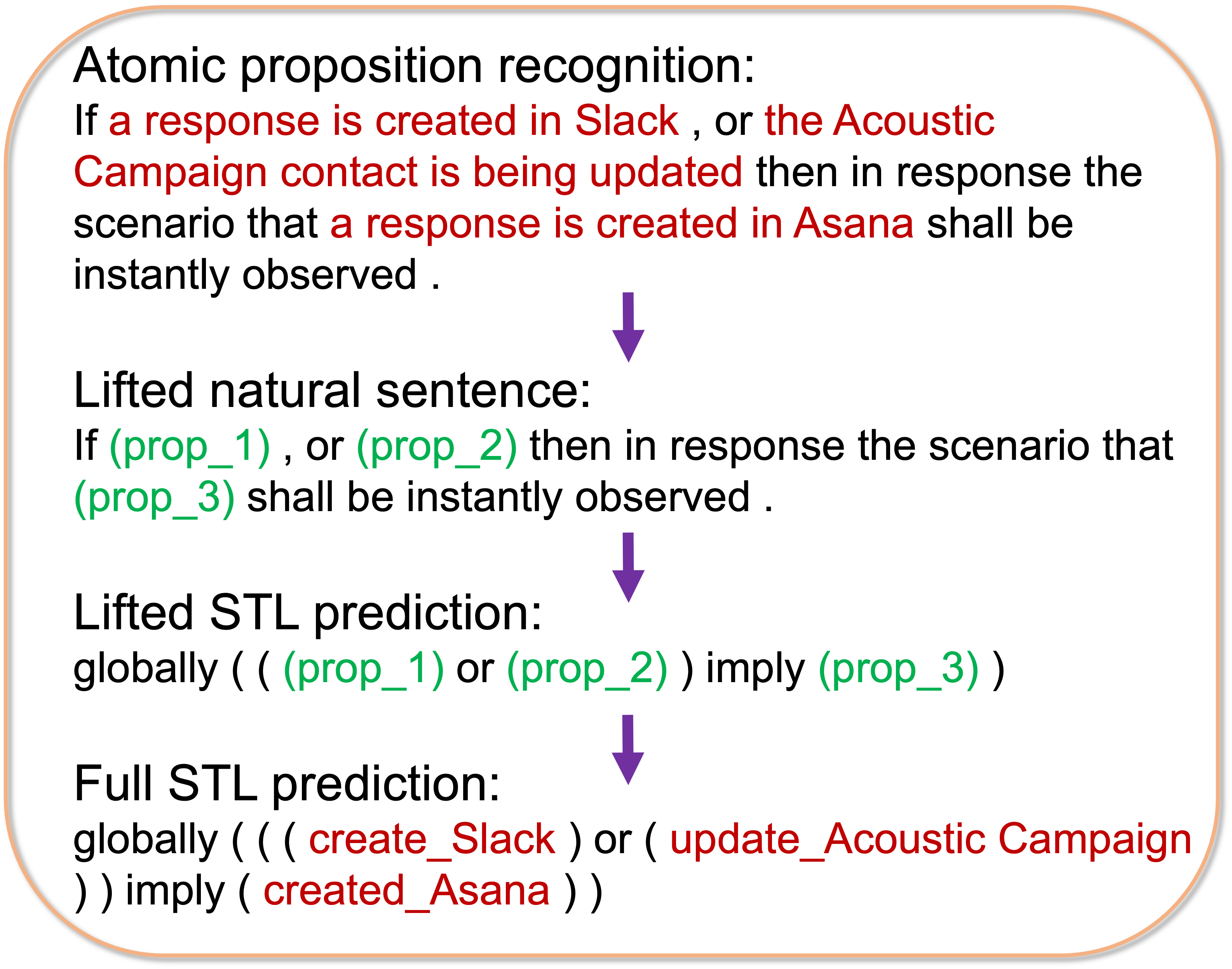}
   \caption{
Illustration of full STL conversion by combing with AP recognition task using GPT-3.}
   \label{fig:AP recognition}
\end{figure}

Illustrated in Figure~\ref{fig:AP recognition}

\end{document}